\documentclass[10pt,twocolumn,letterpaper]{article}

\usepackage[pagenumbers]{cvpr} 

\usepackage{graphicx}
\usepackage{amsmath}
\usepackage{amssymb}
\usepackage{booktabs}
\usepackage{bbm}
\usepackage{url}
\usepackage{multirow,makecell, rotating}
\usepackage{booktabs}
\usepackage{caption}
\usepackage{graphbox}
\usepackage{comment}
\usepackage{epsfig}
\usepackage{bm}
\usepackage{soul}
\usepackage{color}
\usepackage{xcolor}

\usepackage{appendix}

\usepackage[pagebackref,breaklinks,colorlinks]{hyperref}

\usepackage[capitalize]{cleveref}
\crefname{section}{Sec.}{Secs.}
\Crefname{section}{Section}{Sections}
\Crefname{table}{Table}{Tables}
\crefname{table}{Tab.}{Tabs.}

\newcommand*{\affaddr}[1]{#1} 
\newcommand*{\affmark}[1][*]{\textsuperscript{#1}}

\usepackage[symbol]{footmisc}

\begin{document}

\title{Dreamix: Video Diffusion Models are General Video Editors}

\author{%
Eyal Molad\begin{NoHyper}\footnote{}\end{NoHyper}~~\affmark[1], Eliahu Horwitz\begin{NoHyper}\footref{note1}~~\footref{note3}\end{NoHyper}~~\affmark[1,]\affmark[2], Dani Valevski\begin{NoHyper}\footref{note1}\end{NoHyper}~~\affmark[1], Alex Rav Acha\affmark[1], Yossi Matias\affmark[1], Yael Pritch\affmark[1], \\ Yaniv Leviathan \begin{NoHyper}\footref{note2}\end{NoHyper}~~\affmark[1],   Yedid Hoshen\begin{NoHyper}\footref{note2}~~\footref{note3}\end{NoHyper}~~\affmark[1,]\affmark[2]\\
\affaddr{\affmark[1]Google Research,~~}\affaddr{\affmark[2]The Hebrew University of Jerusalem}\\
\small\url{https://dreamix-video-editing.github.io/}
}

\twocolumn[{
	\maketitle
	\vspace{-3em}
	\renewcommand\twocolumn[1][]{#1}
		\begin{center}
		\centering
    \includegraphics[width=0.99\textwidth]{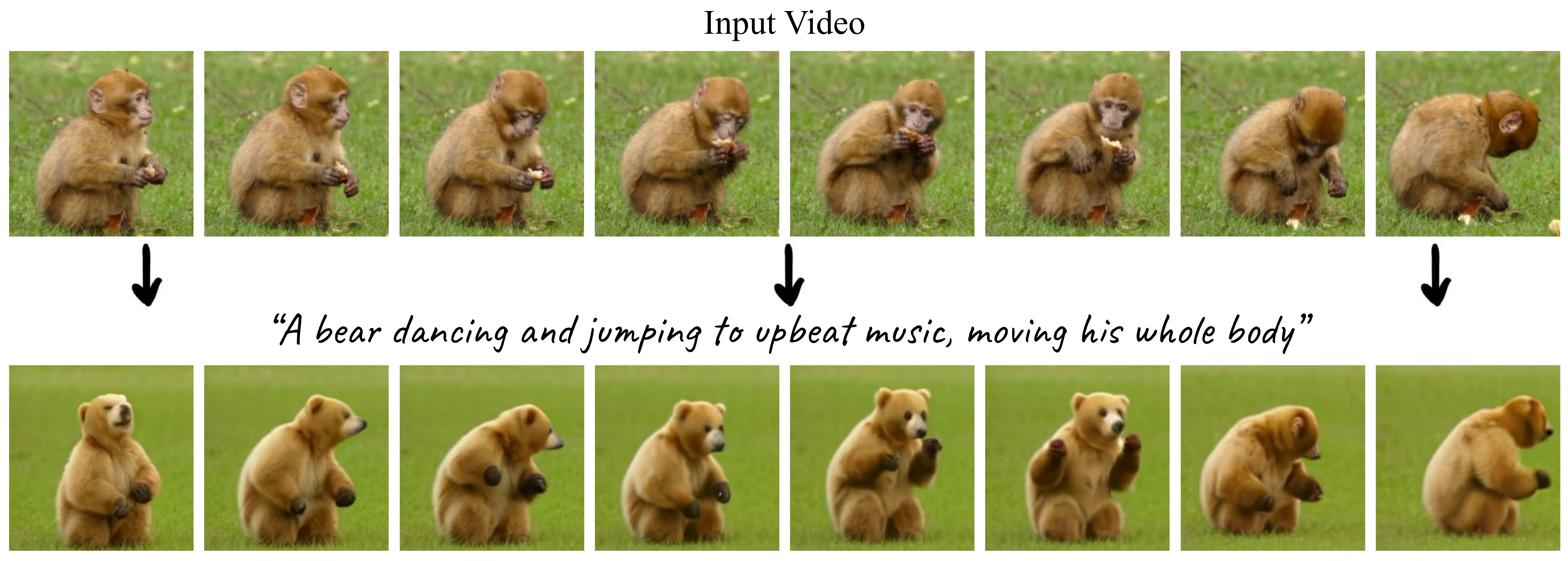}
    \vspace{-0.2cm}
    \captionof{figure}{
    \textbf{\textit{Video Editing with Dreamix}}: Frames from a video conditioned on the text prompt ``A bear dancing and jumping to upbeat music, moving his whole body``. Dreamix transforms the eating monkey (top row) into a dancing bear, affecting appearance and motion (bottom row). It maintains fidelity to color, posture, object size and camera pose, resulting in a temporally consistent video
    }
    \label{fig:teaser}
	\end{center}
}]

\begin{abstract}

Text-driven image and video diffusion models have recently achieved unprecedented generation realism. While diffusion models have been successfully applied for image editing, very few works have done so for video editing. We present the first diffusion-based method that is able to perform text-based motion and appearance editing of general videos. Our approach uses a video diffusion model to combine, at inference time, the low-resolution spatio-temporal information from the original video with new, high resolution information that it synthesized to align with the guiding text prompt. As obtaining high-fidelity to the original video requires retaining some of its high-resolution information, we add a preliminary stage of finetuning the model on the original video, significantly boosting fidelity. We propose to improve motion editability by a new, mixed objective that jointly finetunes with full temporal attention and with temporal attention masking. We further introduce a new framework for image animation. We first transform the image into a coarse video by simple image processing operations such as replication and perspective geometric projections, and then use our general video editor to animate it. As a further application, we can use our method for subject-driven video generation. Extensive qualitative and numerical experiments showcase the remarkable editing ability of our method and establish its superior performance compared to baseline methods.

\end{abstract}

\footnotetext[1]{\label{note1} Equal contribution.}
\footnotetext[2]{\label{note2} Equal advising.}
\footnotetext[3]{\label{note3} Performed this work while working at Google.}

\section{Introduction}
\label{sec:intro}
Recent advancements in generative models \cite{DDPM, maskgit, parti} and multimodal vision-language models \cite{clip} have paved the way to large-scale text-to-image models capable of unprecedented generation realism and diversity \cite{dalle2, ldm, imagen, glide, spatext}. These models have ushered in a new era of creativity, applications, and research efforts. Although these models offer new creative processes, they are limited to synthesizing \textit{new} images rather than editing \textit{existing} ones. To bridge this gap, intuitive text-based image editing methods offer text-based editing of generated and real images while maintaining some of their original attributes \cite{prompt2prompt, pnp, instructpix2pix, imagic, unitune}. Similarly to images, text-to-video models have recently been proposed \cite{video-dm, imagen-video, make-a-video, magvit}, but there are currently very few methods using them for video editing. 

\begin{figure*}[t!]
    \centering
    \includegraphics[width=0.99\linewidth]{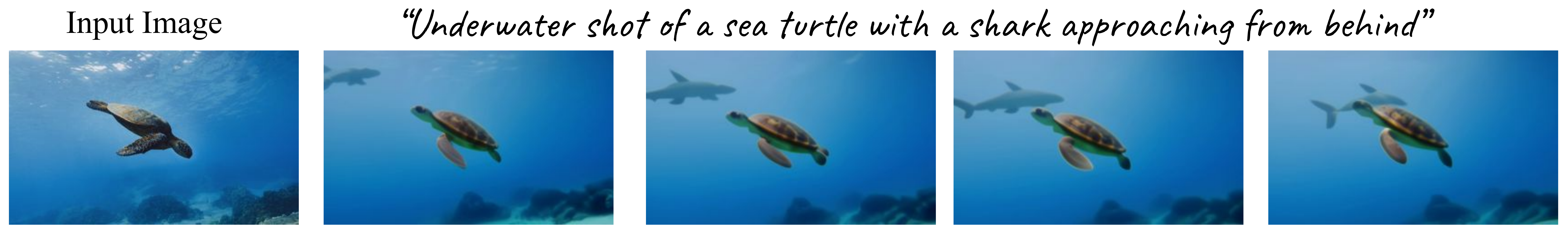}
    \includegraphics[width=0.99\linewidth]{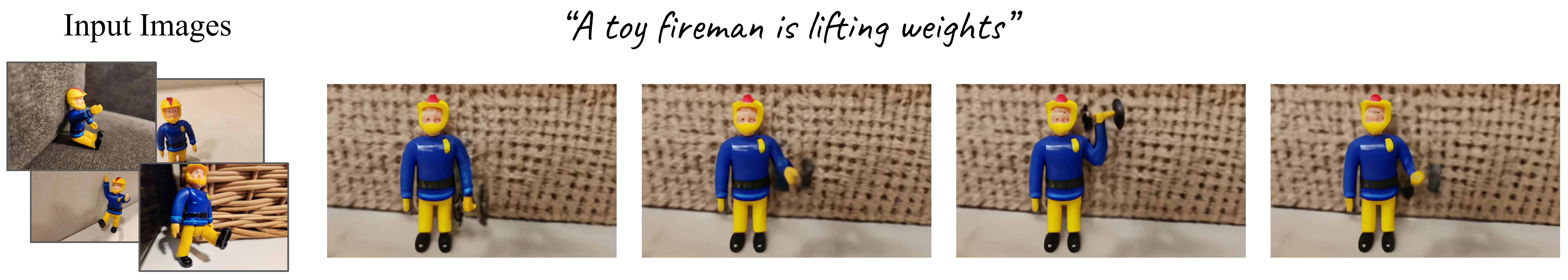}
    
     \caption{\textit{\textbf{Image-to-Video editing with Dreamix:}} Dreamix can create videos based on image and text inputs. In the single image case (first row) it is able to instill complex motion in a static image, adding a moving shark and making the turtle swim. In this case, visual fidelity to object location and background was preserved but the turtle direction was flipped. In the subject-driven case (second row) Dreamix is able to extract the visual features of a subject given multiple images and then animate it in different scenarios such as weightlifting}
    \label{fig:turtle_weights}
\end{figure*}

In text-guided video editing, the user provides an input video and a text prompt which describes the desired attributes of the resulting video (\cref{fig:teaser}). The objectives are three-fold: i) alignment: the edited video should conform with the input text prompt ii) fidelity: the edited video should preserve the content of the original input iii) quality: the edited video should be of high-quality.  Video editing is more challenging than standard image editing, as it requires synthesizing new motion, not merely modifying visual appearance. It also requires temporal consistency. As a result, applying image-level editing methods e.g. SDEdit \cite{sdedit} or Prompt-to-Prompt \cite{prompt2prompt} sequentially on the video frames is insufficient.

We present a new method, Dreamix, to adapt a text-conditioned video diffusion model (VDM) for video editing, in a manner inspired by UniTune \cite{unitune}. The core of our method is enabling a text-conditioned VDM to maintain high fidelity to an input video via two main ideas. First, instead of using pure-noise as initialization for the model, we use a degraded version of the original video, keeping only low spatio-temporal information by downscaling it and adding noise. Second, 
we further improve the fidelity to the original video by finetuning the generation model on the original video. Finetuning ensures the model has knowledge of the high-resolution attributes of the original video.
A naive finetuning on the input video results in relatively low motion editabilty as the model learns to prefer the original motion instead of following the text prompt. We propose a novel, mixed finetuning approach, in which the VDMs are also finetuned on the collection of individual frames of the input video while discarding their temporal order. Technically, this is achieved by masking the temporal attention. Mixed finetuning significantly improves the quality of motion edits. 

As a further contribution, we leverage our video editing model to propose a new framework for image animation (see \cref{fig:turtle_weights}). This has several applications including: animating the objects and background in an image, creating dynamic camera motion, etc. We do this by simple image processing operations, e.g. frame replication or geometric image transformation, to create a coarse video. We then edit it with our Dreamix video editor. We also use our novel finetuning approach for subject-driven video generation, i.e. a video version of Dreambooth\cite{dreambooth}. We perform an extensive qualitative study and a human evaluation, showcasing the remarkable abilities of our method. We compare our method against the state-of-the-art baselines, demonstrating superior results. 
To summarize, our main contributions are:

\begin{enumerate}
    \item Proposing the first method for general text-based appearance and motion editing of real-world videos.
    \item Proposing a novel mixed finetuning model that significantly improves the quality of motion edits.
    \item Presenting a new framework for text-guided image animation, by applying our video editor method on top of simple image preprocessing operations.
    \item Demonstrating subject-driven video generation from a collection of images, leveraging our novel finetuning method.
\end{enumerate}

\begin{figure}[t]
\begin{center}
	\centering
        \includegraphics[width=1\linewidth]{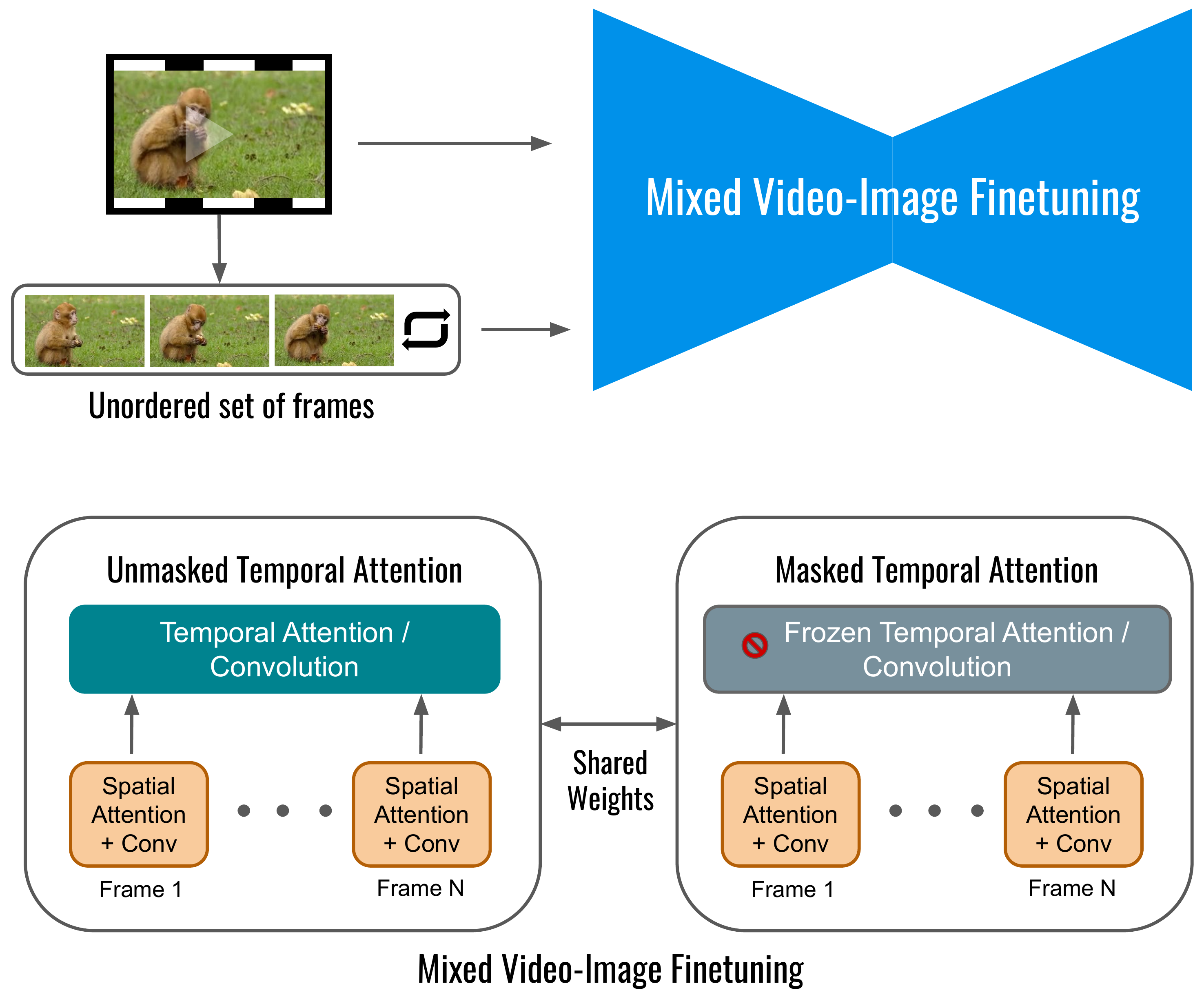}
    \end{center}
 \caption{\textit{\textbf{Mixed Video-Image Finetuning:}} 
 Finetuning the VDM on the input video alone limits the extent of motion change. Instead, we use a mixed objective that beside the original objective (bottom left) also finetunes on the unordered set of frames. This is done by using “masked temporal attention”, preventing the temporal attention and convolution from being finetuned (bottom right). This allows adding motion to a static video}
\label{fig:architecture_finetune}
\end{figure}

\begin{figure*}[t]
    \begin{center}
	\centering
        \includegraphics[width=0.9\linewidth]{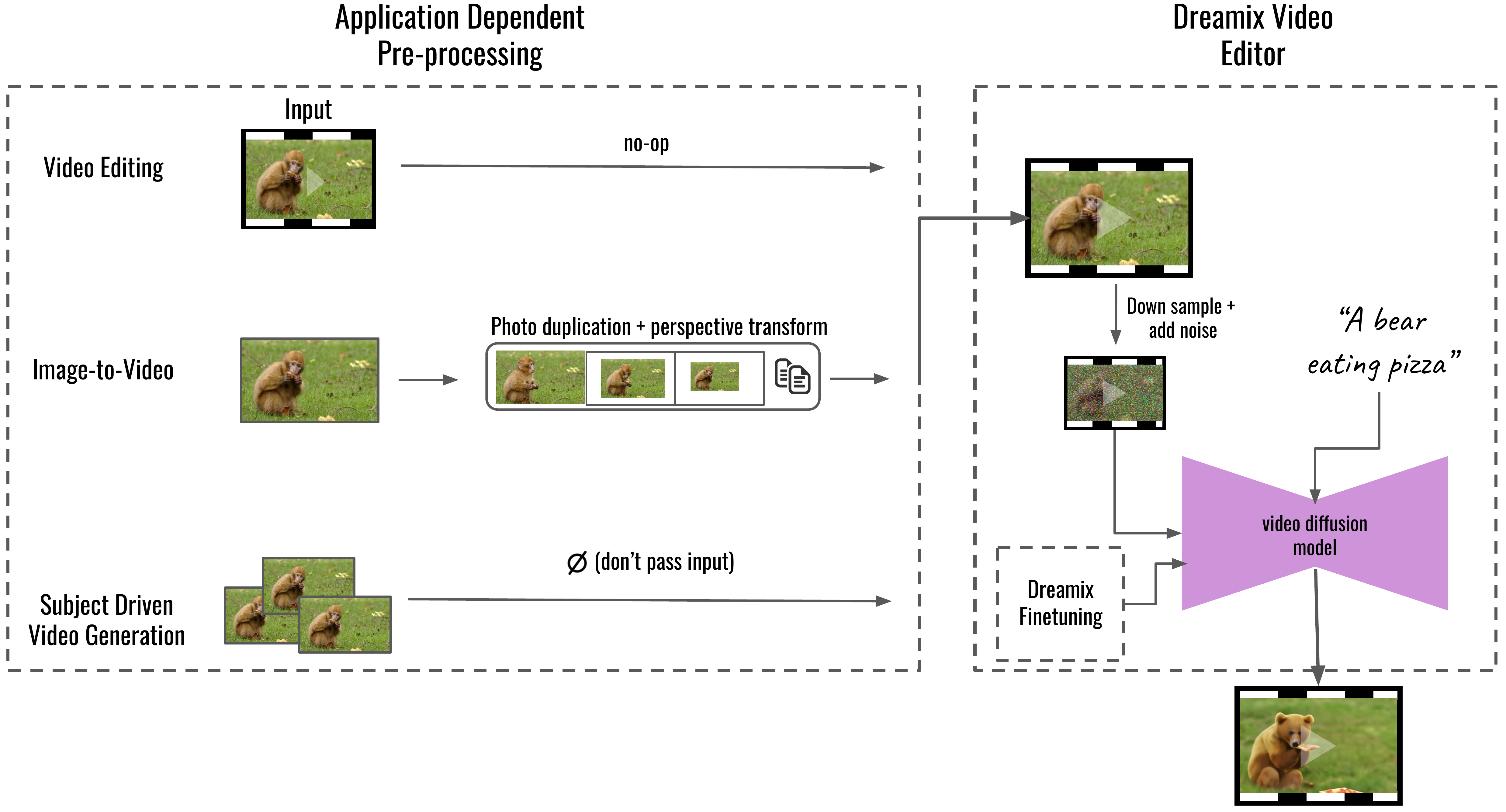}
    \end{center}
 \caption{\textit{\textbf{Inference Overview:}} Our method supports multiple applications by application dependent pre-processing (left), converting the input content into a uniform video format. For image-to-video, the input image is duplicated and transformed using perspective transformations, synthesizing a coarse video with some camera motion. For subject-driven video generation, the input is omitted -  finetuning alone takes care of the fidelity.
 This coarse video is then edited using our general ``Dreamix Video Editor`` (right): we first corrupt the video by downsampling followed by adding noise. We then apply the finetuned text-guided VDM, which upscales the video to the final spatio-temporal resolution}
\label{fig:architecture_applications}
\end{figure*}

\section{Related Work}
\subsection{Diffusion Models for Synthesis}
\label{subsec:related-dm}
Deep diffusion models recently emerged as a powerful new paradigm for image generation \cite{DDPM, DDIM}, and have their roots in score-matching \cite{hyvarinen2005estimation,vincent2011connection,score_matching}. They outperform \cite{ADM} the previous state-of-the-art approach, generative adversarial networks (GANs) \cite{GAN}. While they have multiple formulations, EDM \cite{EDM} showed they are equivalent.  Outstanding progress was made in text-to-image generation \cite{imagen, dalle2, ldm, spatext}, where new images are sampled conditioned on an input text prompt. Extending diffusion models to video generation is a challenging computational and algorithmic task.
Early work include \cite{video-dm} and text-to-video extensions by \cite{imagen-video, make-a-video}.  Another line of work extends synthesis to various image reconstruction tasks \cite{sr3, palette, cascaded, repaint, come_closer}, \cite{conffusion} extracts confidence intervals for reconstruction tasks.

\subsection{Diffusion Models for Editing}
\label{subsec:related-editing}
Image editing with generative models has been studied extensively, in past years many of the models were based on GANs\cite{deepsim, styleclip, stylegan-nada, pix2pixhd, spade}. Editing methods have recently adopted diffusion models \cite{blended_diffusion, blended_latent_diffusion, sketch-dm}. Several works proposed to use text-to-image diffusion models for editing rather than text-conditioned synthesis. SDEdit \cite{sdedit} proposed to add targeted noise and other corruptions to an input image, and then use diffusion models for reversing the process. It can perform significant image edits, while losing some fidelity to the original image. Prompt-to-Prompt \cite{prompt2prompt} (and later Plug-and-Play \cite{pnp} and \cite{null-text-inv}) perform semantic edits by mixing activations extracted with the original and target prompts. For InstructPix2Pix \cite{instructpix2pix} this is only needed at test time. Other works (e.g. \cite{text_inv,dreambooth}) use finetuning and optimization to allow for personalization of the model, learning a special token describing the content. UniTune \cite{unitune} and Imagic \cite{imagic} finetune on a single image, allowing better editability while maintaining good fidelity. However, the above methods are image-centric and do not take temporal information into account. Text2Live \cite{text2live} allows some texture-based video editing but are not diffusion-based and cannot edit motion. A concurrent paper, Tune-a-Video \cite{tune-a-video} preform video editing by inflating a text-to-image model to learn temporal consistency. Despite their promising results, they use a text-to-image backbone that can edit video appearance but not motion. Their results are also not fully temporally consistent. In contrast, our method uses a text-to-video backbone, enabling motion editing while maintaining video smoothness.

\section{Background: Video Diffusion Models}
\label{sec:background}

\textbf{Denoising Model Training.} Diffusion models rely on deep denoising neural network $D_{\theta}$. Let us denote the groundtruth video as $v$, an i.i.d Gaussian noise tensor of the same dimensions as the video as $\epsilon \sim N(0, \textbf{I})$, and the noise level at time $s$ as $\sigma_s$. The noisy video is given by: $z_s = \gamma_s v + \sigma_s \epsilon$, where $\gamma_s = \sqrt{1 - \sigma^2_s}$. Furthermore, let us denote a conditioning text prompt as $t$ and a conditioning video $c$ (for super-resolution, $c$ is a low-resolution version of $v$). The objective of the denoising network $D_{\theta}$ is to recover the groundtruth video $v$ given the noisy input video $z_s$, the time $s$, prompt $t$ and conditioning video $c$. The model is trained on a (typically very large) training corpus $\mathcal{V}$ consisting of pairs of video $v$ and text prompts $t$. The optimization objective is:

\begin{equation}
\mathcal{L}_{\theta}(v) = \mathbb{E}_{\epsilon \sim N(0, \textbf{I}), s \in \mathcal{U}(0,1)}  \| D_{\theta}(z_s, s, t, c) - v\|^2
\end{equation}

\textbf{Sampling from Diffusion Models.} The key challenge in diffusion models is to use the denoiser network $D$ to sample from the distribution of videos conditioned on the text prompt $t$ and conditioning video $c$, $P(v|t,c)$. While the derivation of such sampling rule is non-trivial (see e.g. \cite{EDM}), the implementation of such sampling is relatively simple in practice. We follow \cite{imagen-video} in using stochastic DDIM sampling. At a heuristic level, at each step, we first use the densoier network to estimate the noise. We then remove a fraction of the estimated noise and finally add randomly generated Gaussian noise, with magnitude corresponding to half of the removed noise.   

\textbf{Cascaded Video Diffusion Models.} Training high-resolution text-to-video models is very challenging due to the high computational complexity. Several diffusion models overcome this using cascaded architectures. We use Imagen-Video \cite{imagen-video}, which consists of a cascade of $7$ models. The base model maps the input text prompt into a ~$5$-second video of $24 \times 40 \times 16$ frames. It is then followed by $3$ spatial super-resolution models and $3$ temporal super-resolution models. For implementation details, see \cref{app:implementation-details}.

\section{General Editing by Video Diffusion Models}
\label{sec:method}

We propose a new method for video editing using text-guided video diffusion models. We extended it to image animation in \cref{sec:applications}.

\subsection{Text-Guided Video Editing by Inverting Corruptions}
\label{subsec:inference}

We wish to edit an input video using the guidance of a text prompt $t$ describing the video \textbf{after} the edit. In order to do so we leverage the power of a cascade of VDMs. The key idea is to first corrupt the video by downsampling followed by adding noise. We then apply the sampling process of the cascaded diffusion models from the time step corresponding to the noise level, conditioned on $t$, which upscales the video to the final spatio-temporal resolution.
The effect is that the VDM will use the low-resolution details provided by the degraded input video, but synthesize new high spatio-temporal resolution information using the text prompt guidance.
While this procedure is essentially a text-guided version of SDEdit \cite{sdedit}, there are some video specific technical challenges that we will describe below. Note, that this by itself does not result in sufficiently high-fidelity video editing. We present a novel finetuning objective for mitigating this issue in \cref{subsec:finetuning}. 

\textbf{Input Video Degradation.} 
We downsample the input video to the resolution of the base model ($16$ frames of $24 \times 40$). We then add i.i.d Gaussian noise with variance $\sigma_{s}^2$ to further corrupt the input video. The noise strength is equivalent to time $s$ in the diffusion process of the base text-to-video model. For $s=0$, no noise is added, while for $s=1$, the video is replaced by pure Gaussian noise. Note that even when no noise is added, the input video is highly corrupted due to the extreme downsampling ratio. For the non-finetuned base model, values of $s \in [0.4,0.85]$ typically worked best.

\begin{figure*}[t!]
    \centering
        \includegraphics[width=0.99\linewidth]{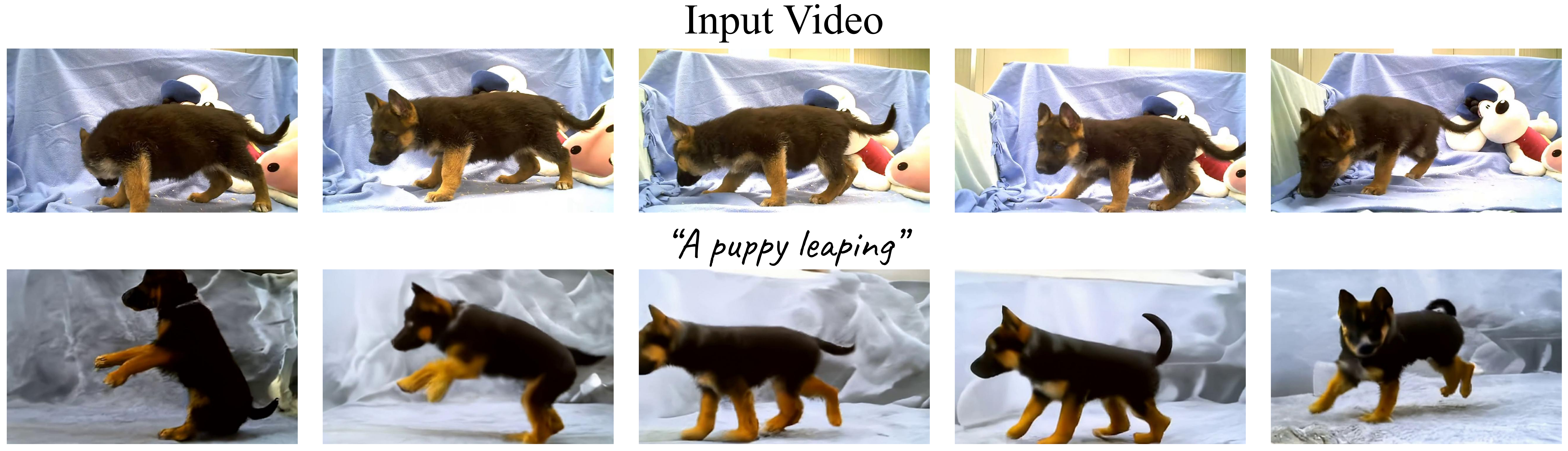}
     \caption{\textit{\textbf{Video Motion Editing:}} Dreamix can significantly change the actions and motions of subjects in a video, making a puppy leap in this example. The resulting video maintains temporal consistency while preserving the unedited details}
    \label{fig:leapingdog}
\end{figure*}

\textbf{Text-Guided Corrpution Inversion.} 
We can now use the cascaded VDMs  to map the corrputed, low-resolution video into a high-resolution video that aligns with the text.  The core idea here is that given a noisy, very low spatio-temporal resolution video, there are many perfectly feasible, high-resolution videos that correspond to it. We use the target text prompt $t$ to select the feasible outputs that not only correspond to the low-resolution of the original video but are also aligned to edits desired by the user.  The base model starts with the corrupted video, which has the same noise as the diffusion process at time $s$. We use the model to reverse the diffusion process up to time $0$. We then upscale the video through the entire cascade of super-resolution models (see \cref{app:implementation-details}). All models are conditioned on the prompt $t$.

\subsection{Mixed Video-Image Finetuning}
\label{subsec:finetuning}

The naive method presented in \cref{subsec:inference} relies on a corrupted version of the input video which does not include enough information to preserve high-resolution details such as fine textures or object identity. We tackle this issue by adding a preliminary stage of finetuning the model on the input video $v$. Note that this only needs to be done once for the video, which can then be edited by many prompts without furher finetuning. We would like the model to separately update its prior both on the appearance and the motion of the input video. Our approach therefore treats the input video, both as a single video clip and as an unordered set of $M$ frames, denoted by $u = \{x_1,x_2,..,x_M\}$. We use a rare string $t^*$ as the text prompt, following \cite{dreambooth}. We finetune the denoising models by a combination of two objectives. The first objective updates the model prior on both motion and appearance by requiring it to exactly reconstruct the input video $v$ given its noisy versions $z_s$.

\begin{equation}
\mathcal{L}^{vid}_{\theta}(v) = \mathbb{E}_{\epsilon \sim N(0, \textbf{I}), s \in \mathcal{U}(0,1)}  \| D_{\theta'}(z_s, s, t^*, c) - v\|^2
\end{equation}

Additionally, we train the model to reconstruct each of the frames individually given their noisy version. This enhances the appearance prior of the model, separately from the motion. Technically, the model is trained on a sequence of frames $u$ by replacing the temporal attention layers by trivial fixed masks ensuring the model only pays attention within each frame, and also by masking the residual temporal convolution blocks. We denote the attention masked denoising model as $D^a_{\theta}$. The masked attention objective is given by: 
\begin{equation}
\mathcal{L}^{frame}_{\theta}(u) = \mathbb{E}_{\epsilon \sim N(0, \textbf{I}), s \in \mathcal{U}(0,1)}  \| D^a_{\theta'}(z_s, s, t^*, c) - u\|^2
\end{equation}

\noindent We train the objectives jointly and denote this \textit{mixed finetuning}:
\begin{equation}
\theta = arg\min_{\theta'} \alpha \mathcal{L}^{vid}_{\theta'}(v) + (1-\alpha) \mathcal{L}^{frame}_{\theta'}(u)
\end{equation}
Where $\alpha$ is a hyperparameter weighting between the two objectives, (see \cref{fig:architecture_finetune}).  Training on a single video or a handful of frames can easily lead to overfitting, reducing the editing ability of the original model. To mitigate overfitting, we use a small number of finetuning iterations and a low learning rate (see \cref{app:implementation-details}).

\subsection{Hyperparameters}
\label{subsec:hyperparams}

\begin{figure}[t]
    \includegraphics[width=0.99\linewidth]{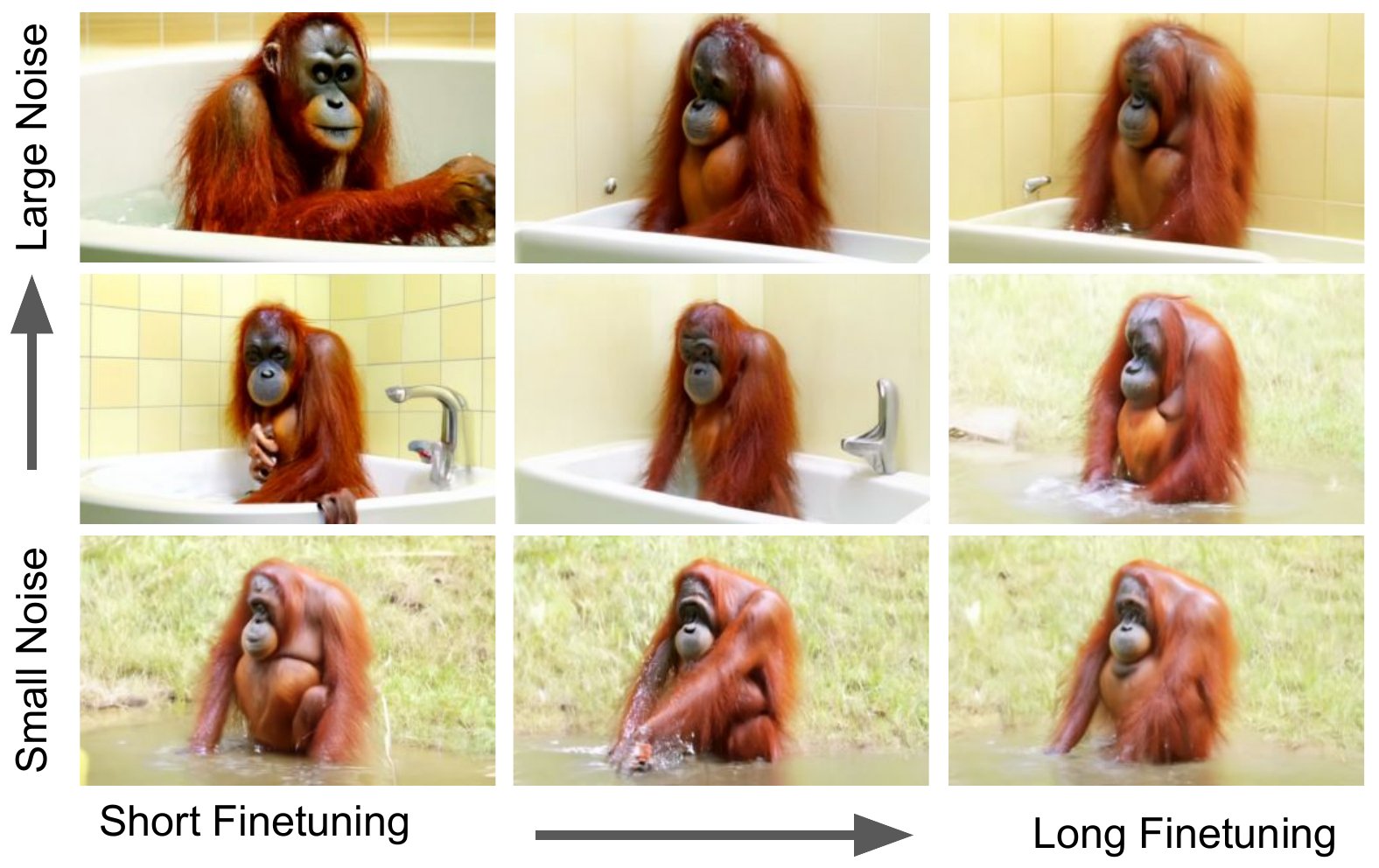}
 \caption{
 \textit{\textbf{Hyperparameter tradeoffs:}}  We compare the effect of noise magnitude and number of finetuning iterations on edited videos. The original frame is on the bottom left. The rest are frames generated by different parameters for the prompt "An orangutan with orange hair bathing in a bathroom". We can observe that higher noise allows for larger edits but reduces fidelity. More finetuning iterations improve fidelity at higher noises. The best results are obtained for high noise and a large number of finetuning iterations 
 }
\label{fig:ablation}
\end{figure}

Our method has several hyperparameters. For inference time, we have the noise scale $s \in [0,1]$ where $s=1$ corresponds to standard sampling without using the degraded input video. For finetuning, we have the number of finetuning steps $FT_{steps}$, learning rate $lr$, and mixing weight $\alpha$ between the video and frames finetuning objectives (see \cref{subsec:finetuning}). 
See \cref{fig:ablation} for a qualitative analysis of hyperparameter impact, and \cref{subsec:ablation} for a quantatitve analysis. Additional implementation details may be found in \cref{app:implementation-details}.

\begin{figure*}[t!]
    \centering
    \includegraphics[width=0.99\linewidth]{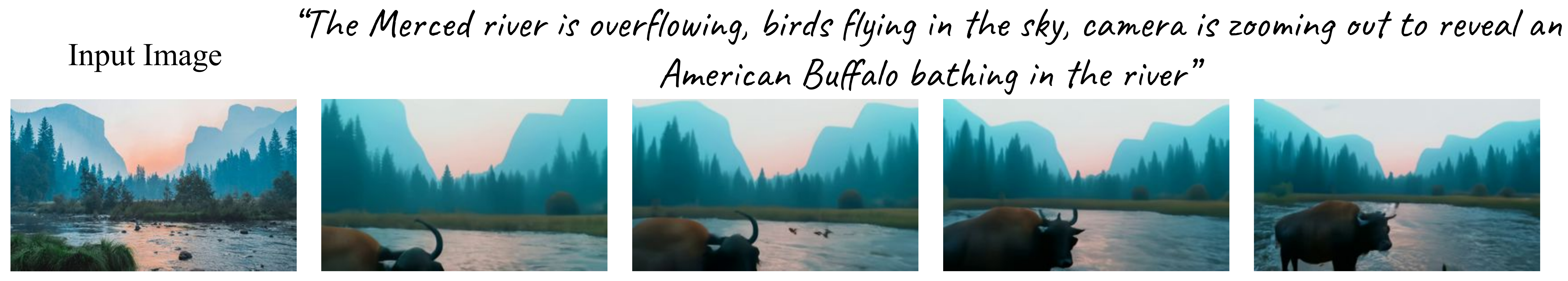}
        \includegraphics[width=0.99\linewidth]{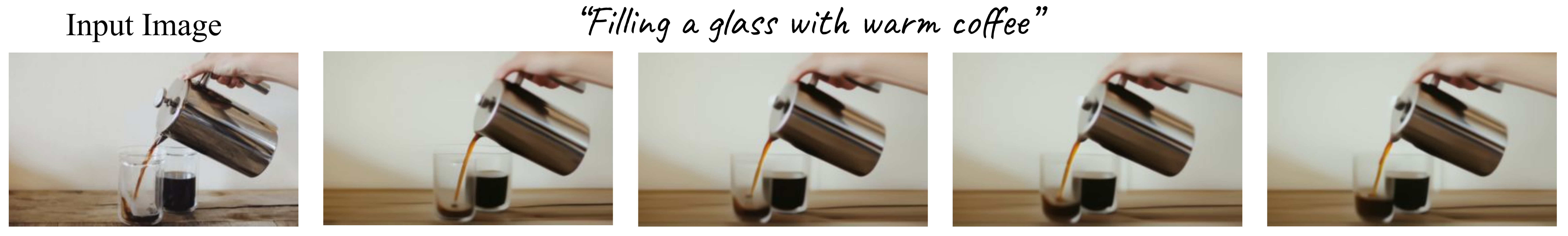}
    \includegraphics[width=0.99\linewidth]{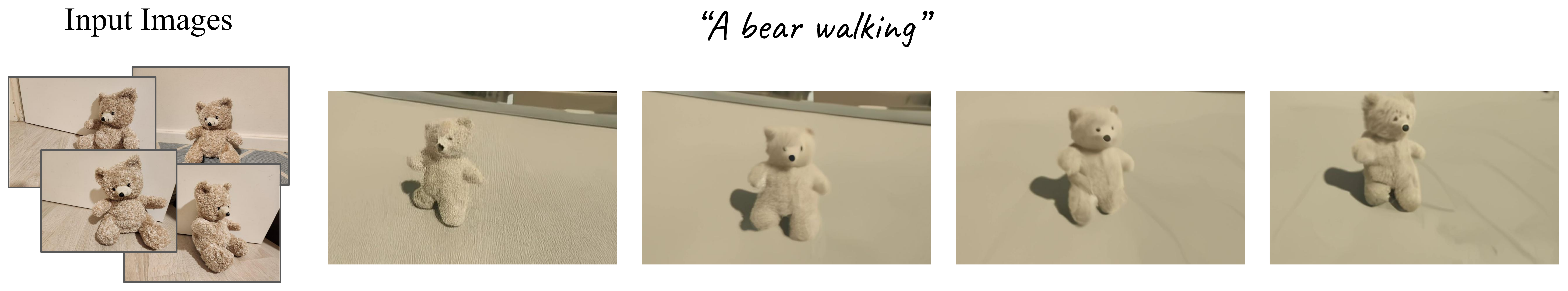}
     \caption{\textit{\textbf{Additional Image-to-Video Results:}} Dreamix can generate camera effects such as zoom-out by combining a text prompt with a coarse video obtained by applying image-processing transformations on the input image. First row - the image is zoomed out to reveal a bathing buffalo. Dreamix can also instill motion in a static image as in the second row where the glass is gradually filled with coffee. Third row - we animate a provided subject based on a small number of independent images}
    \label{fig:yosemite_coffee_bear}
\end{figure*}

\section{Applications of Dreamix}
\label{sec:applications}

The method proposed in \cref{sec:method}, can naturally be used to edit motion and appearance in real-world videos. In this section, we propose a framework for using our Dreamix video editor for general, text-conditioned image-to-video editing, see \cref{fig:architecture_applications} for an overview.

\textbf{Dreamix for Single Images.} Provided our general video editing method, Dreamix, we now propose a framework for image animation conditioned on a text prompt. The idea is to transform the image or a set of images into a coarse, corrupted video and edit it using Dreamix. For example, given a single image $x$ as input, we can transform it to a video by replicating it $16$ times to form a static video $v = [x,x,x...x]$. We can then edit its appearance and motion using Dreamix conditioned on a text prompt. Here, we do not wish to incorporate the motion of the input video (as it is static and meaningless) and therefore use only the masked temporal attention finetuning ($\alpha=0$). We can further control the output video, by simulating camera motion, such as panning and zoom. We perform this by sampling a smooth sequence of $16$ perspective transformations $T_1,T_2..T_{16}$ and apply each on the original image. When the perspective requires pixels outside the input image, we simply outpaint them using reflection padding. We concatenate the sequence of transformed images into a low quality input video $v=[T_1(x),T_2(x)..T_{16}(x)]$. While this does not result in realistic video, Dreamix can transform it into a high-quality edited video.  

\textbf{Dreamix for subject-driven video generation.} We propose to use Dreamix for text-conditioned video generation given an image collection. The input to our method is a set of images, each containing the subject of interest. This can potentially also use different frames from the same video, as long as they show the same subject. Higher diversity of viewing angles and backgrounds is beneficial for the performance of the method. We then use our novel finetuning method from \cref{subsec:finetuning}, where we only use the masked attention finetuning ($\alpha=0$). After finetuning, we use the text-to-image model \textit{without} a conditioning video, but rather only using a text prompt (which includes the special token $t^*$).   
\section{Experiments}

\subsection{Qualitative Results}

We showcase the results of Dreamix, demonstrating unprecendented video editing and image animation abilities. 

\begin{figure*}[t]
    \centering
    \includegraphics[width=0.95\linewidth]{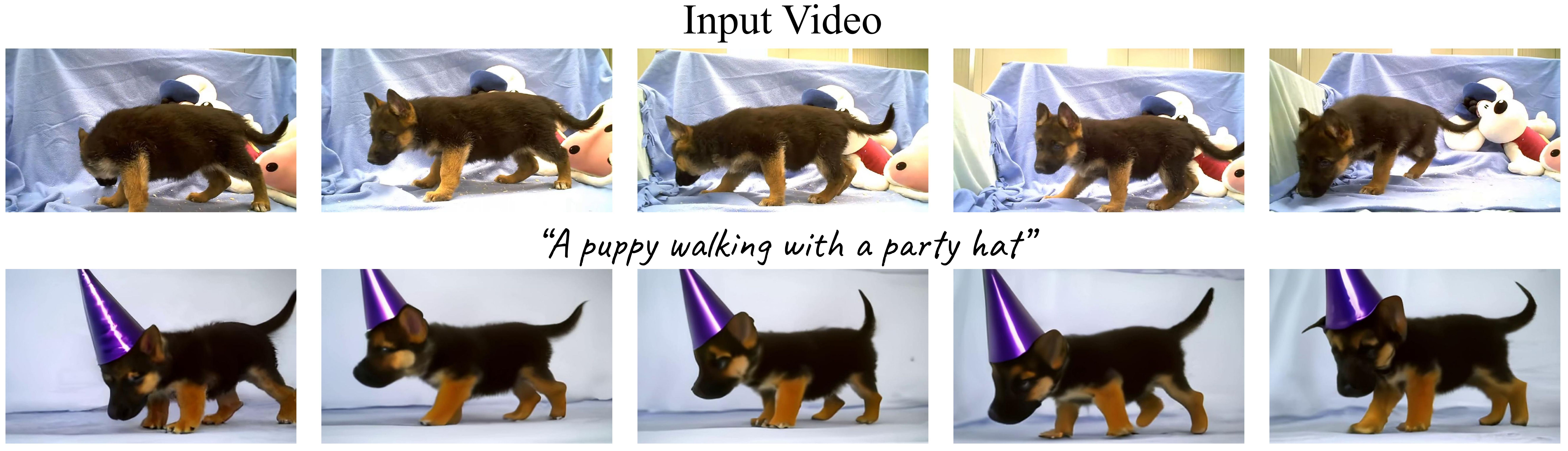}
   \includegraphics[width=0.95\linewidth]{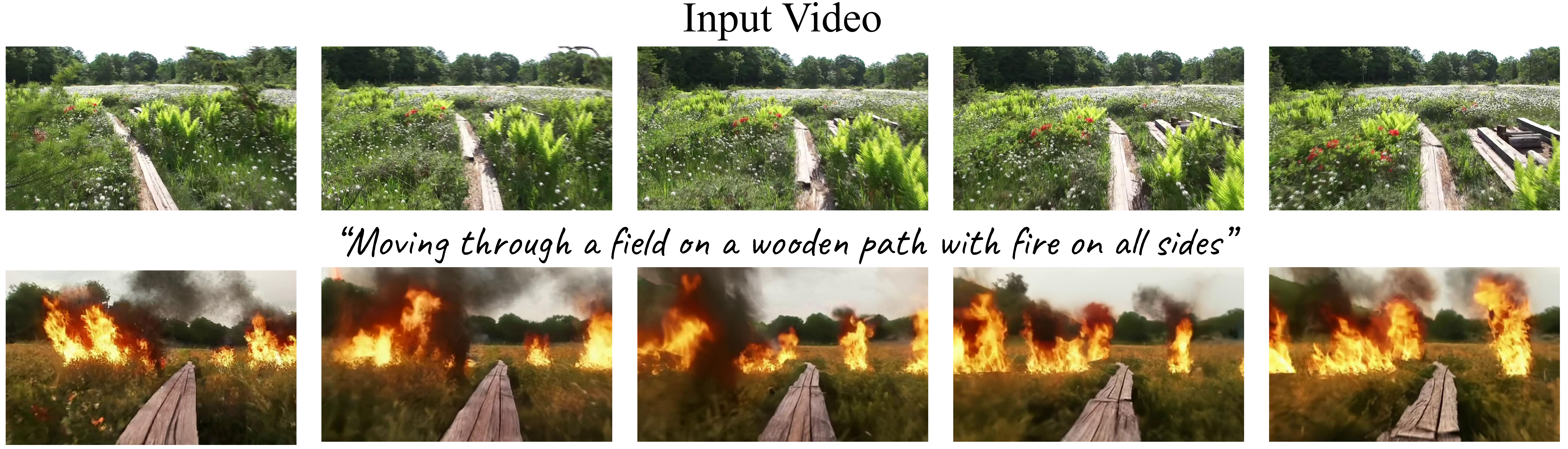}
     \includegraphics[width=0.95\linewidth]{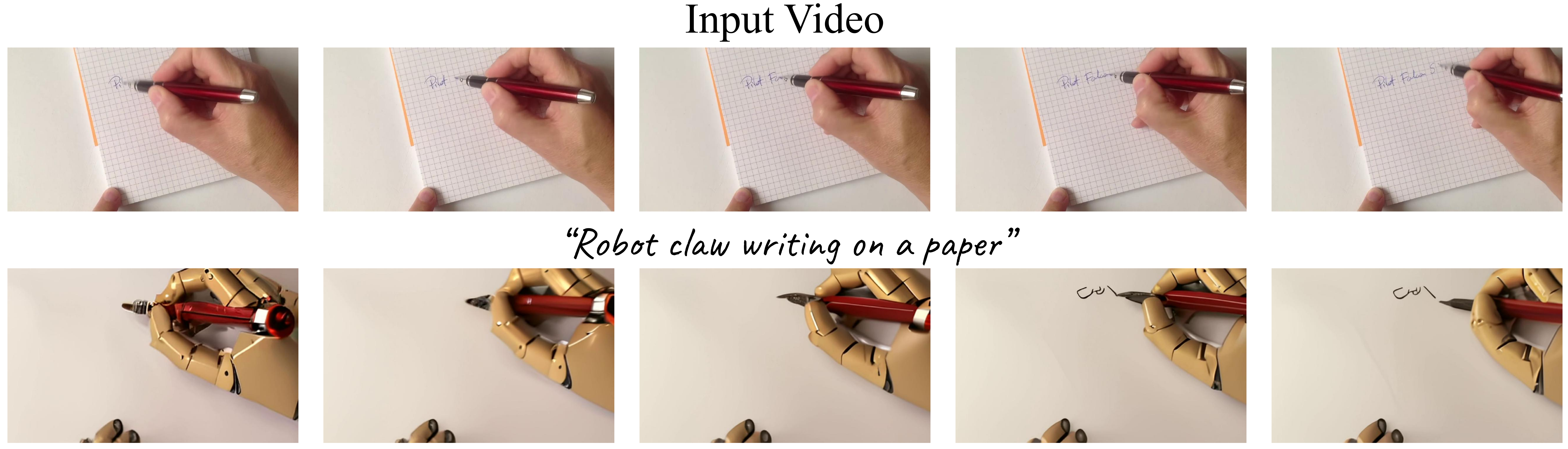}  
     \caption{\textit{\textbf{Additional Video-to-Video results}} }
    \label{fig:hat_field_pen}
\end{figure*}

\textbf{Video Editing.} In \cref{fig:teaser}, we change the motion to dancing and the appearance from monkey to bear. keeping the coarse attributes of the video fixed. Dreamix can also generate new motion that does not necessarily align with the input video (puppy in \cref{fig:leapingdog}, orangutan in \cref{fig:sm_video_editing3}), and can control camera movements (zoom-out example in \cref{fig:sm_video_editing4}). Dreamix can generate smooth visual modifications that align with the temporal information in the input video. This includes adding effects (field in \cref{fig:hat_field_pen}, saxophone in the \cref{fig:sm_video_editing4}), adding objects (hat in \cref{fig:hat_field_pen} and skateboard in \cref{fig:sm_video_editing1}) or replacing them (robot in \cref{fig:hat_field_pen}), changing the background (truck in the \cref{fig:sm_video_editing4}).

\begin{figure*}
    \centering
    \includegraphics[width=0.95\linewidth]{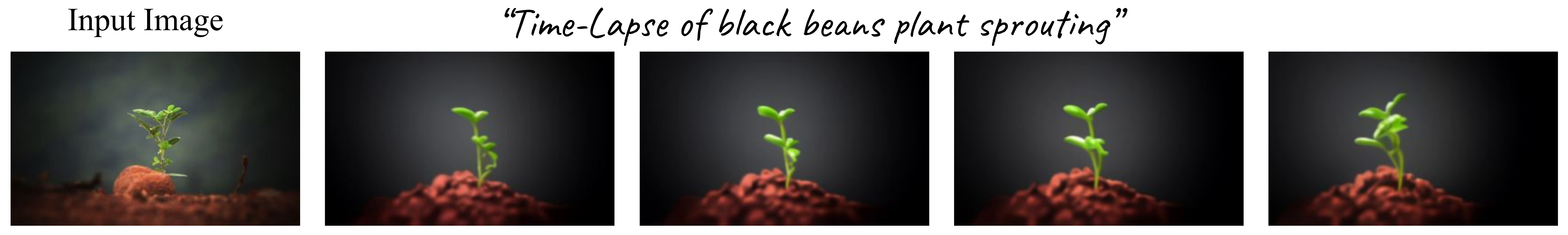}
    \includegraphics[width=0.95\linewidth]{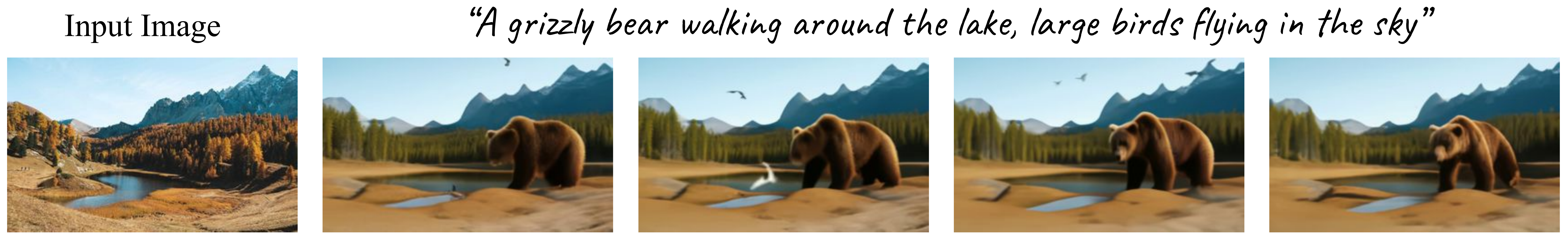}
    \includegraphics[width=0.95\linewidth]{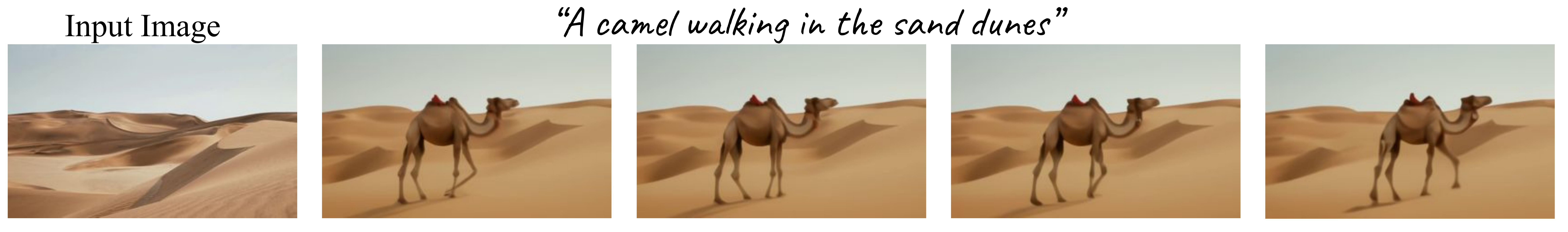}
    \includegraphics[width=0.95\linewidth]{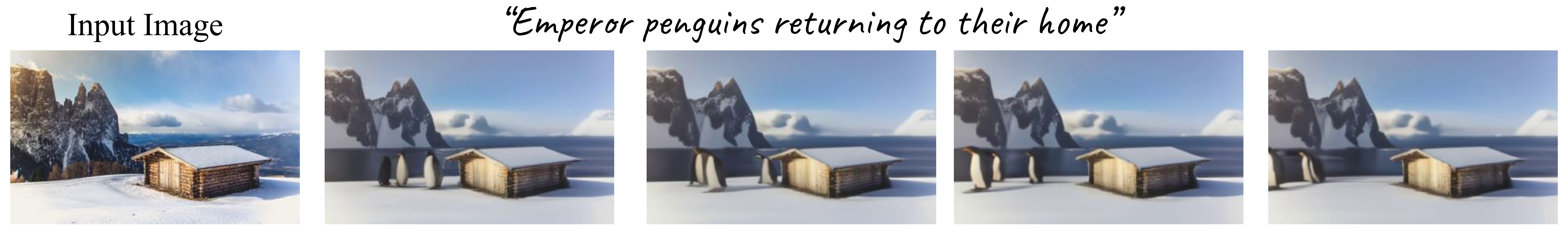}
    \includegraphics[width=0.95\linewidth]{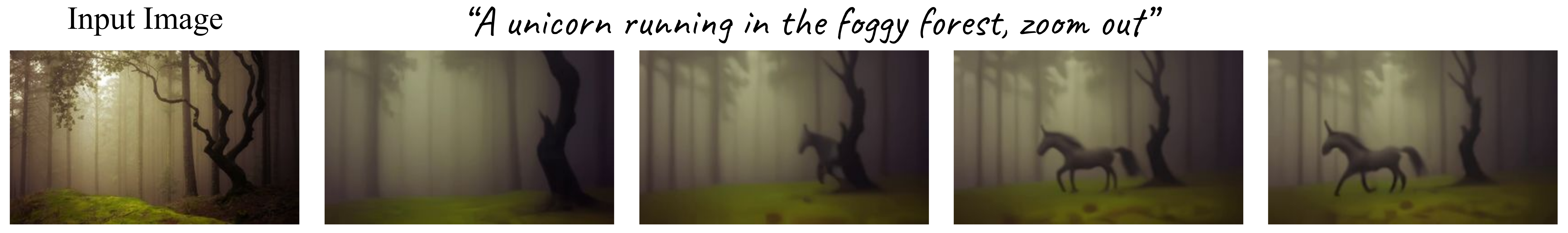}    
    \includegraphics[width=0.95\linewidth]{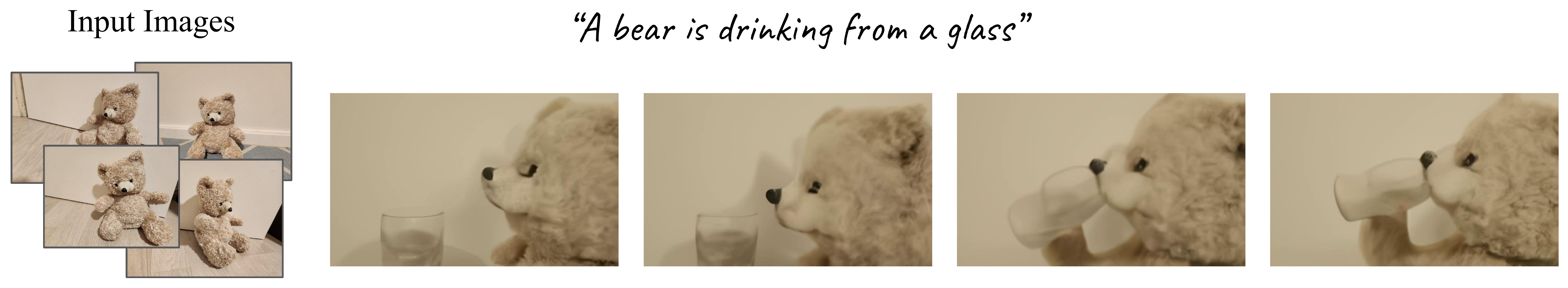}
    \includegraphics[width=0.95\linewidth]{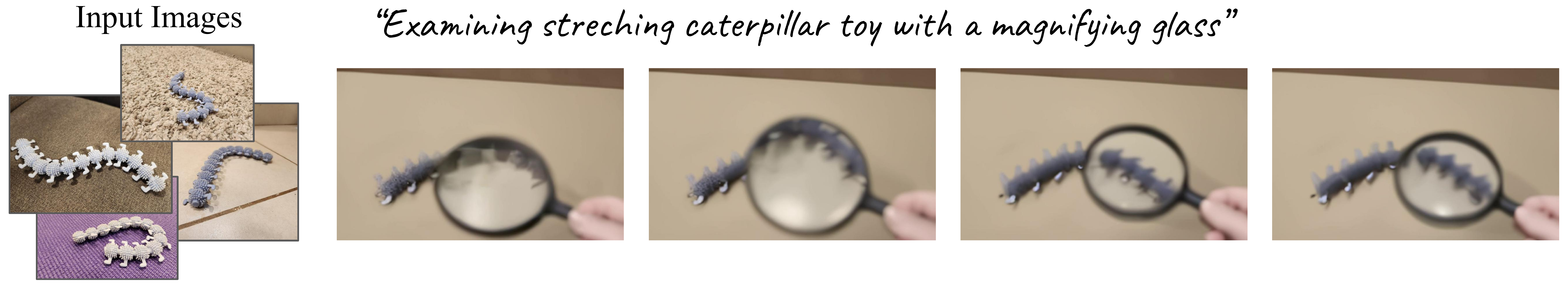}
    \includegraphics[width=0.95\linewidth]{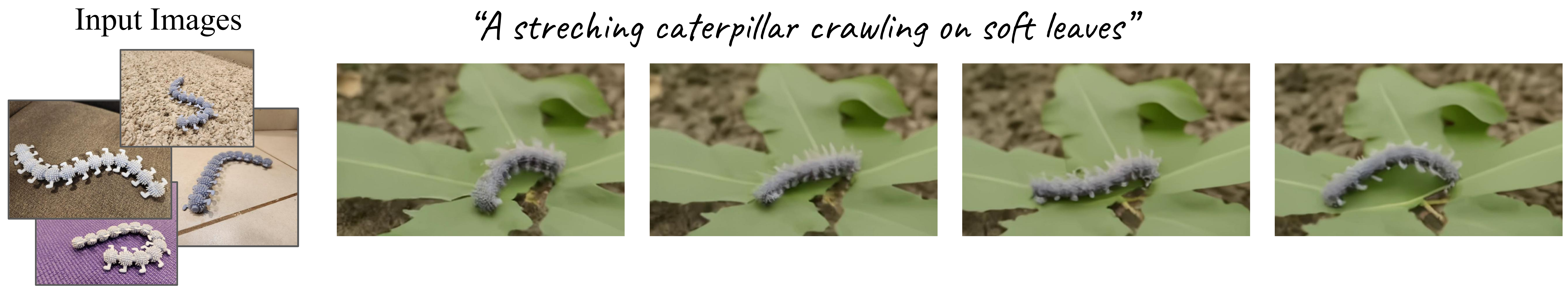}
     \caption{\textit{\textbf{Additional Results:}} Image-to-Video, and subject-driven video generation}
    \label{fig:grizzly_camel_plant_penguin_unicorb_drinking_magnifying_stretching}
\end{figure*}

\textbf{Image-driven Videos.} When the input is a single image, Dreamix can use its video prior to to add new moving objects (camel in \cref{fig:grizzly_camel_plant_penguin_unicorb_drinking_magnifying_stretching}), inject motion into the input (turtle in \cref{fig:turtle_weights} and coffee in \cref{fig:yosemite_coffee_bear}), or create new camera movements (buffalo in \cref{fig:yosemite_coffee_bear}). Our method is unique in being able to do this for general, real-world images.

\begin{table}[t]

\caption{\textbf{Ablation Study:} Users were asked to compare text-guided video edits of different variants of our method: no finetuning (no ft.), video-only finetuning (Video ft.), the proposed mixed finetuning (Mixed-ft). The object category includes adding/replacing objects. The background category includes background, color or texture changes. Mixed finetuning is important in motion editing and background change scenarios}
\begin{center}
    \begin{tabular}{l@{\hskip5pt}c@{\hskip5pt}c@{\hskip5pt}c@{\hskip5pt}c@{\hskip5pt}c@{\hskip5pt}}
    
    Edit Type  & \# Edits & No ft.  & Video-ft.  & Mixed-ft. & None   \\
     \midrule     
     Motion & 36  & 17\% & 25\% & 47\% & 11\%  \\   
     Style  & 15  & 67\% & 7\% & 20\% & 6\%  \\
     Object & 44  & 36\% &	30\% &	18\% & 16\% \\
     Background & 32 & 19\% & 28\% & 44\% & 9\% \\
     \bottomrule
    \end{tabular}
\end{center}

\label{table:method_human_pref}
\end{table}

\textbf{Subject-driven Video Generation.} Dreamix can take an image collection showing the same subject and generate new videos with this subject in motion. This is unique, as previous approaches could only do this for images. We demonstrate this on a range of subjects and actions including: the weight-lifting toy fireman in \cref{fig:turtle_weights}, walking and drinking bear in \cref{fig:yosemite_coffee_bear} and \cref{fig:grizzly_camel_plant_penguin_unicorb_drinking_magnifying_stretching}. It can place the subjects in new surroundings, e.g., moving caterpillar to a leaf in \cref{fig:grizzly_camel_plant_penguin_unicorb_drinking_magnifying_stretching} and even under a magnifying glass in \cref{fig:grizzly_camel_plant_penguin_unicorb_drinking_magnifying_stretching}.

\begin{table}[t]
\caption{\textbf{Baseline Comparison:} Users were asked to rate videos edited by different methods by visual quality, fidelity to the base video and alignment with the text prompt. We define an edit as successful when it receives a mean score larger than 2 in all dimensions. We observe that our method significantly outperforms the others in producing successful edits}
\begin{center}
    \begin{tabular}{l@{\hskip5pt}c@{\hskip5pt}c@{\hskip5pt}c@{\hskip5pt}c@{\hskip5pt}}    
    Method  & Quality  & Fidelity  & Alignment & Success   \\
     \midrule

    ImgenVid & $2.99$ {\textbf{\scriptsize\textcolor{gray}{$\pm$0.95}}} & $2.21$ {\textbf{\scriptsize\textcolor{gray}{$\pm$0.97}}} & $4.04$ {\textbf{\scriptsize\textcolor{gray}{$\pm$1.12}}} & $40\%$    \\   
  
     PnP  & $1.76$ {\textbf{\scriptsize\textcolor{gray}{$\pm$0.78}}}  & $3.61$ {\textbf{\scriptsize\textcolor{gray}{$\pm$0.96}}} & $3.09$ {\textbf{\scriptsize\textcolor{gray}{$\pm$1.35}}} & $15\%$  \\
     
     Ours & $3.09$ {\textbf{\scriptsize\textcolor{gray}{$\pm$0.83}}} & $3.29$ {\textbf{\scriptsize\textcolor{gray}{$\pm$0.99}}} & $3.50$ {\textbf{\scriptsize\textcolor{gray}{$\pm$1.34}}} & $73\%$  \\     
     \bottomrule
    \end{tabular}
\end{center}

\label{table:baseline_comp}
\end{table}

\begin{figure}[t]
    \includegraphics[width=0.99\linewidth]{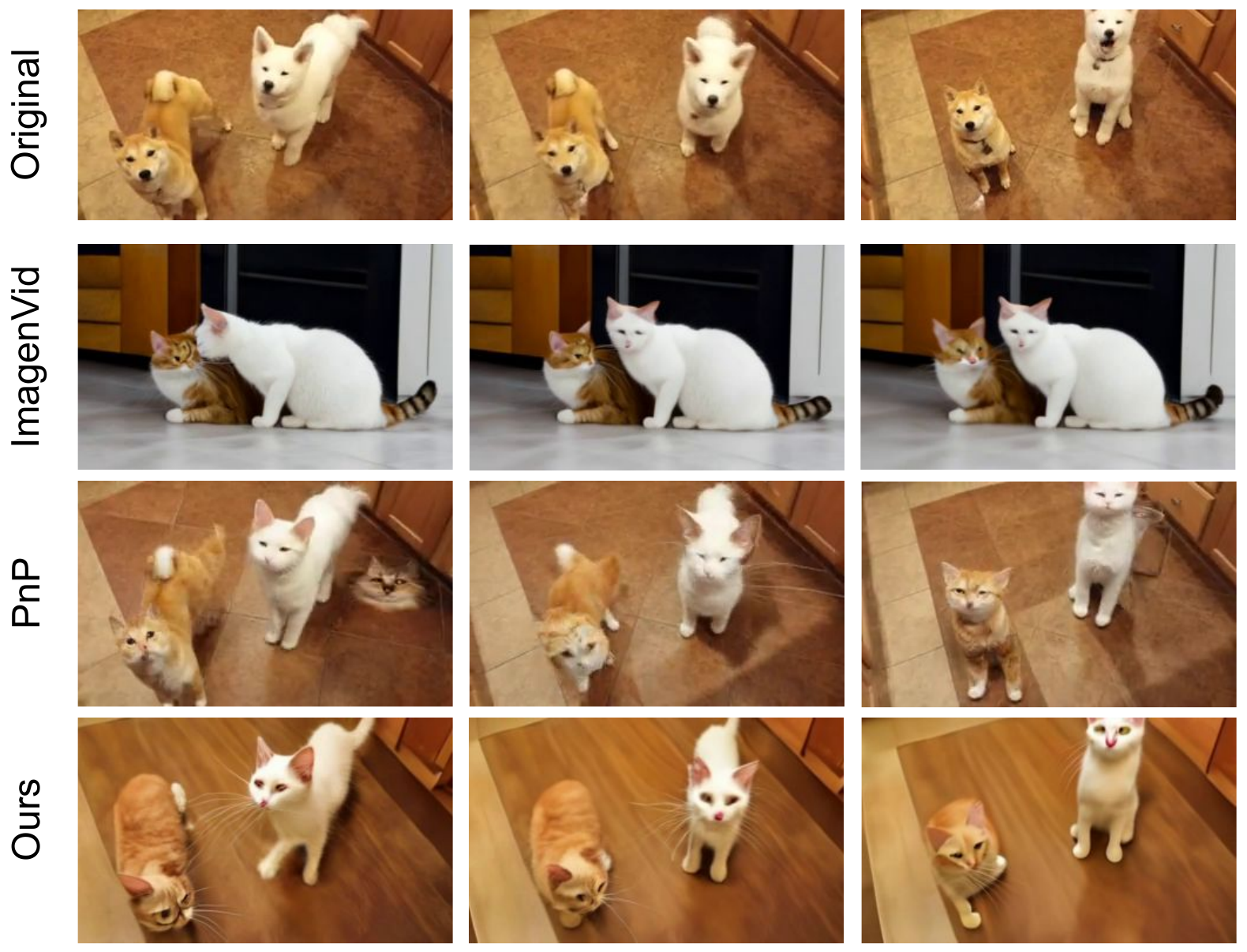}
 \caption{
 \textit{\textbf{Comparison to Baseline Methods:}} Top row - original video. Second row - Imagen-Video. Although the quality and text alignment are high, there is no resemblance to our original video. Third row - PnP, an image-based text-guided editing method. Although the scene is well preserved (e.g. tiles), the temporal consistency is low. Bottom - ours. The edit aligns well with the text prompt while preserving many original details and generating a high quality video
 }
\label{fig:comparisons}
\end{figure}

\subsection{Baseline Comparisons}
\label{sec:comparisons}

\textbf{Baselines.} We compare our method against two baselines:

\textit{Text-to-Video.} Directly mapping the text prompt to a video, without conditioning on the input video using Imagen-Video.

\textit{Plug-and-Play (PnP).} A common approach for video editing is to apply text-to-image editing on each frame individually. We apply PnP\cite{pnp} (a SoTA method) on each frame independently and concatenate the frames into a video.

\textbf{Quantitative Comparison.} We performed a human-rated evaluation of Dreamix and the baselines on a dataset of $29$ videos taken from YouTube-8M \cite{Youtube8M}, and $127$ text prompts, across different categories. We used a single hyperparamter set for all methods. Each edited video was rated on a scale of $1-5$ to evaluate its visual quality, its fidelity to the unedited details of the base video and its alignment with the text prompt. We collected $4-6$ ratings for each edited video.
The results of the evaluation can be seen in \cref{table:baseline_comp}. We also highlight the success rate of each method, where a successful edit is one that received a mean score larger than 2 in all dimensions. We observe that frame by frame methods like Plug-and-Play\cite{pnp} perform poorly in terms of visual quality as they create flickering effects due to the lack of temporal input. Moreover, Plug-and-Play sometimes ignored the edit altogether, resulting in low alignment and high fidelity. The Text-to-Video baseline ignores the edited video, resulting in low fidelity. Our method balances between the three dimensions, resulting in a high success rate.

\textbf{Qualitative Comparison.} \cref{fig:comparisons} presents an example of a video edited by Dreamix and the two baselines. The text-to-video model achieves low fidelity edits as it is not conditioned on the original video. PnP preserves the scene but lacks consistency between different frames. Dreamix performs well in all three objectives.

\subsection{Ablation Study}
\label{subsec:ablation}

We conducted a user study comparing our proposed mixed finetuning method (See \cref{subsec:finetuning}) to two ablations: no finetuning and finetuning on the video only (but not the independent frames). Our dataset contained $29$ videos (each of $5$ seconds) taken from YouTube-8M \cite{Youtube8M}, and a total of $127$ text prompts. Additional details are provided in \cref{app:human-eval-details}. 
The results are presented in \cref{table:method_human_pref}. Our main observations are: \textit{Motion} changes require high-editability. Frame-based finetuning typically outperformed video-only finetuning. Denoising without finetuning worked well for \textit{style transfer}, finetuning was often detrimental. Preserving fine-details in \textit{background, color or texture} changes required finetuning.

\section{Discussion}
\label{sec:discussion}

In this section, we analyse the limitations of our method, potential ways to address them and future applications. 

\textbf{Hyperparameter Selection.} Optimal hyperparameter values e.g., noise strength, can change between prompts. Automating their selection will make our method more user friendly. It can be done by learning a regressor from (input video, prompt) to the optimal hyperparameters. Creating a training set with the optimal hyperparameters per-edit (e.g. as judged by users) is left for future work.

\textbf{Automatic Evaluation Metrics.} In our preliminary study, we found that automatic evaluation metrics (e.g. CLIP Score \cite{clipscore} for alignment) are imperfectly correlated with human preference. Future work on automatic video text-editing metrics should address this limitation. Having effective metrics will also support labeling large datasets for the automatic hyperparameter selection suggested above.

\textbf{Frequency of Objects in Dataset and Editability.} Not all prompt-video pairs yield successful edits (as can be seen in \cref{table:baseline_comp}). Being able to determine the successful pairs in advance, will speed up the creative editing process. In preliminary work, we found that edits containing objects and actions that frequently occurred in the training dataset resulted in better edits than rarer ones. This suggests that an automatic method for prompt engineering is a promising direction.

\textbf{Computational Cost.} VDMs are computationally expensive. Finetuning our model, containing billions of parameters, requires large hardware accelerators around $30$ minutes per video. Speeding it up and lowering the computational cost, will allow our method to be used for a larger set of applications.  

\textbf{Future Applications.} We expect Dreamix to have many future applications. Several promising ones are: motion interpolation between an image pair, text-guided inpainting and outpainting.

\section{Conclusion}
\label{sec:conclusion}

We presented a general approach for text-conditioned editing using video diffusion models. Beyond video editing, we introduced a new framework for image animation. We also applied our method to subject-driven video generation. Extensive experiments demonstrated the unprecedented results of our method.

\section{Social Impact}
Our primary aim in this work is to advance research on tools to enable users to animate their personal content. While the development of end-user applications is out of the scope of this work, we recognize both the opportunities and risks that may follow from our contributions. As discussed above, we anticipate multiple possible applications for this work that have the potential to augment and extend creative practices. The personalized component of our approach brings particular promise as it will enable users to better align content with their intent, despite potential biases present in general VDMs. On the other hand, our method carries similar risks as other highly capable media generation approaches. Malicious parties may try to use edited videos to mis-lead viewers or to engage in targeted harassment. Future research must continue investigating these concerns.

\section{Acknowledgements}
We thank Ely Sarig for creating the video, Jay Tenenbaum for the video narration, Amir Hertz for the implementation of our eval baseline, Daniel Cohen-Or, Assaf Zomet, Eyal Segalis, Matan Kalman and Emily Denton for their valuable inputs that helped improve this work.

\small{
\bibliographystyle{ieee_fullname}
\bibliography{refs}
}

\clearpage
\appendix
\appendixpage

\section{Implementation Details}
\label{app:implementation-details}
\subsection{Architecture}
All of our experiments were preformed on Imagen-Video \cite{imagen-video}, a pertrained cascaded video diffusion model, with the following components:
\begin{enumerate}
\item a T5-XXL\cite{t5} text encoder, that computes embeddings from the textual prompt. This embeddings are then used as a condition by all other models.
\item a base video diffusion model, conditioned on text. It generates videos at $16 \times 24 \times 40 \times 3$ resolution (frames $X$ height $X$ width $X$ channels) at $3$ fps.
\item  $6$ super-resolution video diffusion models, each conditioned on text and on the output video of the previous model. Each model is  either spatial (SSR), i.e. upscales resolution, or temporal (TSR), i.e. fills in intermediate frames between the input frames. The order of super resolution models is TSR (2x), SSR (2x), SSR(4x), TSR(2x), TSR(2x), and SSR(4x). The multiplier in the parenthesis for output frames (for TSR), and for output pixels in height and width (for SSR). The final output video is in $128 \times 768 \times 1280 \times 3$ at $24$ fps.
\end{enumerate}

Note that the diffusion models are pretrained on both videos and images, with frozen temporal attention and convolution for the latter. Our mixed finetuning approach treats video frames as if they were images.

\textbf{Distillation.} For some of these models, we use a distilled version to allow for faster sampling times.
The base model is a distilled model with $64$ sampling steps. The first two SSR models are non-distilled models with $128$ sampling steps (due to finetuning considerations, see below). All other SR models use $8$ sampling steps. All models use classifier-free-guidance weight of 1.0 (meaning that classifier free guidance is turned off).

\subsection{Finetuning}
To reduce finetuning time, we only finetune the base model and the first 2 SSR models. In our experiments, finetuning the first 2 SSR models using the distilled models (with $8$ sampling steps) did not yield good quality. We therefore use the non-distilled versions of these models for all experiments (including non-finetuned experiments).
Good combinations of finetuning hyperparameters are:
\begin{itemize}
\item $\alpha = 1.0$ (video only finetuning), $FT_{steps}=64$
\item $\alpha = 0.35$ (mixed video / video-frame finetuning),  $FT_{steps} \in [200, 300]$
\item $\alpha = 0$ (video-frame only finetuning, $FT_{steps} \in [50, 150]$
\end{itemize}
The learning rate ($lr$) we use in all experiments is $6 \cdot 10^{-6}$, much lower then the value used for pretraining the models.

\subsection{Sampling}
We use a DDIM sampler with stochastic noise correction, following \cite{imagen-video}.
For the last highest resolution SSR, for capacity reasons, we use the model to sample a sub-chunks of 32 frames of the input lower resolution videos, and then we concatenate all the outputs together back to 128 frame videos.

\textbf{Noise strength.}
We got the best results for the following values of noise strength $s$: for non-finetuned models, $s \in [0.4, 0.85]$ and for finetuned models, $s \in [0.95, 1.0]$.

\section{Human evaluations details}
\label{app:human-eval-details}
We performed human evaluations for the baseline comparison and the ablation analysis. Both evaluations were conducted by a panel of $10$ human raters, over a dataset of $29$ videos with $127$ edit prompts. The dataset videos were selected from YouTube-8M \cite{Youtube8M} and show animals, people performing actions, vehicles, and other objects. The edit prompt categories are detailed in \cref{table:method_human_pref} of the main paper. The video resolution shown to raters was $350 \times 200$.

In the ablation analysis the raters selected the best edited video out of $12$ hyperparameter combinations.

In the baseline comparison, the raters saw the original video alongside an edited video and answered the following questions:
\begin{enumerate}
\item Rate the overall visual quality and smoothness of the edited video.
\item How well does the edited video match the textual edit description provided?
\item How well does the edited video preserve unedited details of the original video?
\end{enumerate}
 We used a single set of hyperparmeters in the baseline eval: $\alpha=0.35; FT_{steps}=300; s=1$.

\section{Image Attribution}
\label{app:attribution}
\begin{itemize}
\item  Desert - \url{https://unsplash.com/photos/PP8Escz15d8}
\item  Fuji mountain \url{https://unsplash.com/photos/9Qwbfa_RM94}
\item  Tree in snow - \url{https://unsplash.com/photos/aQNy0za7x0k}
\item  Hut in snow - \url{https://unsplash.com/photos/qV2p17GHKbs}
\item  Lake with trees - \url{https://unsplash.com/photos/dIQlgwq6V3Y}
\item  Plant - \url{https://unsplash.com/photos/LrPKL7jOldI}
\item  Turtle - \url{https://unsplash.com/photos/za9MCg787eI}
\item  Yosemite - \url{https://unsplash.com/photos/NRQV-hBF10M}
\item  Foggy forest - \url{https://unsplash.com/photos/pKNqyx_v62s}
\item  Coffee - \url{https://unsplash.com/photos/SMPe5xfbPT0}
\item  Monkey - \url{https://www.pexels.com/video/a-brown-monkey-eating-bread-2436088/}
\end{itemize}

\begin{figure*}[h!]
\begin{tabular}{c@{\hskip1pt}}
\includegraphics[width=0.98\linewidth]{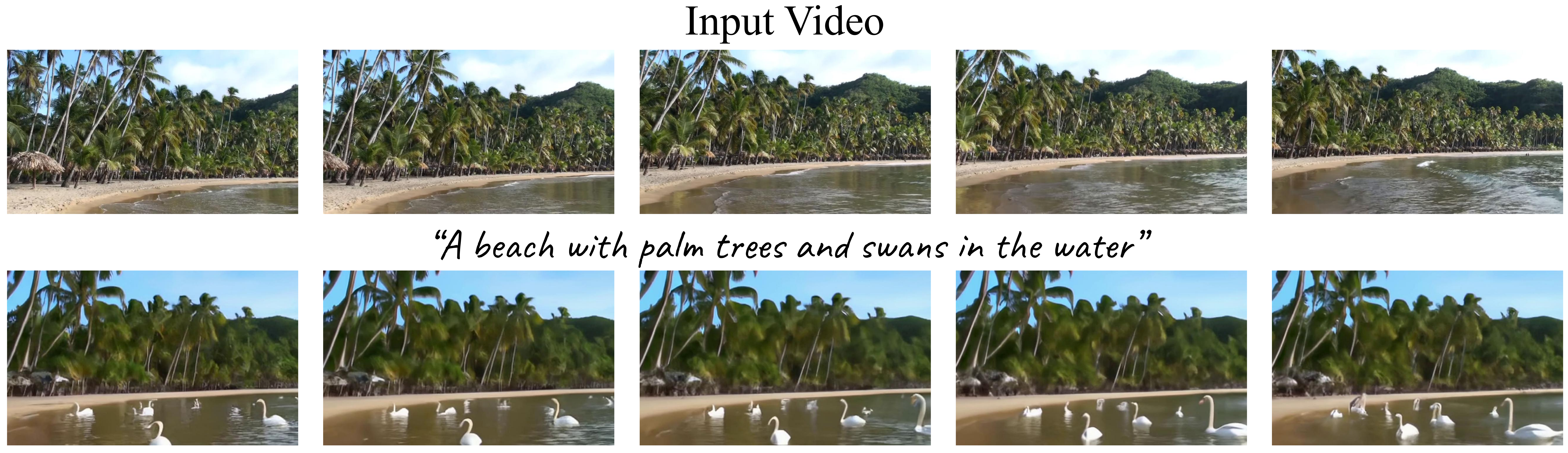} \\
\includegraphics[width=0.98\linewidth]{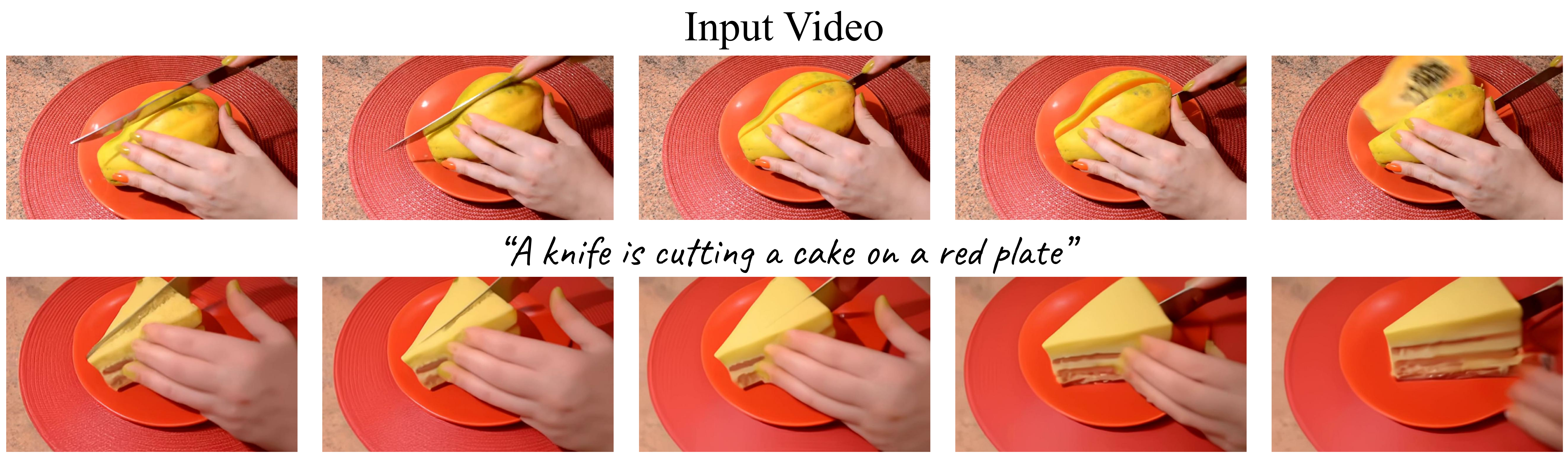}  \\
\includegraphics[width=0.98\linewidth]{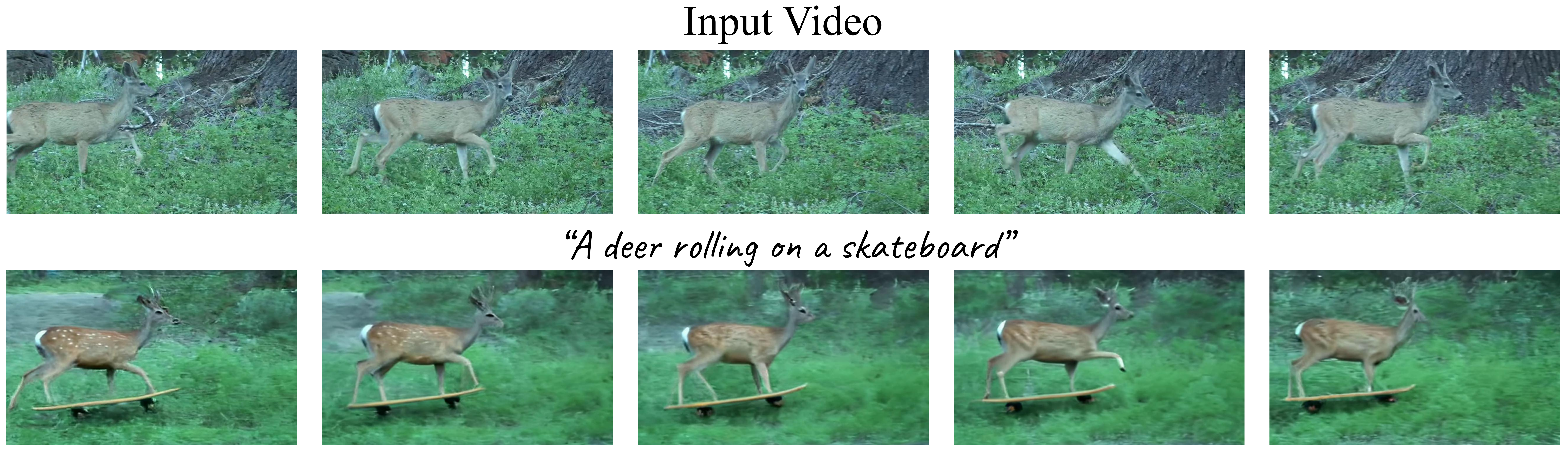} \\
\includegraphics[width=0.98\linewidth]{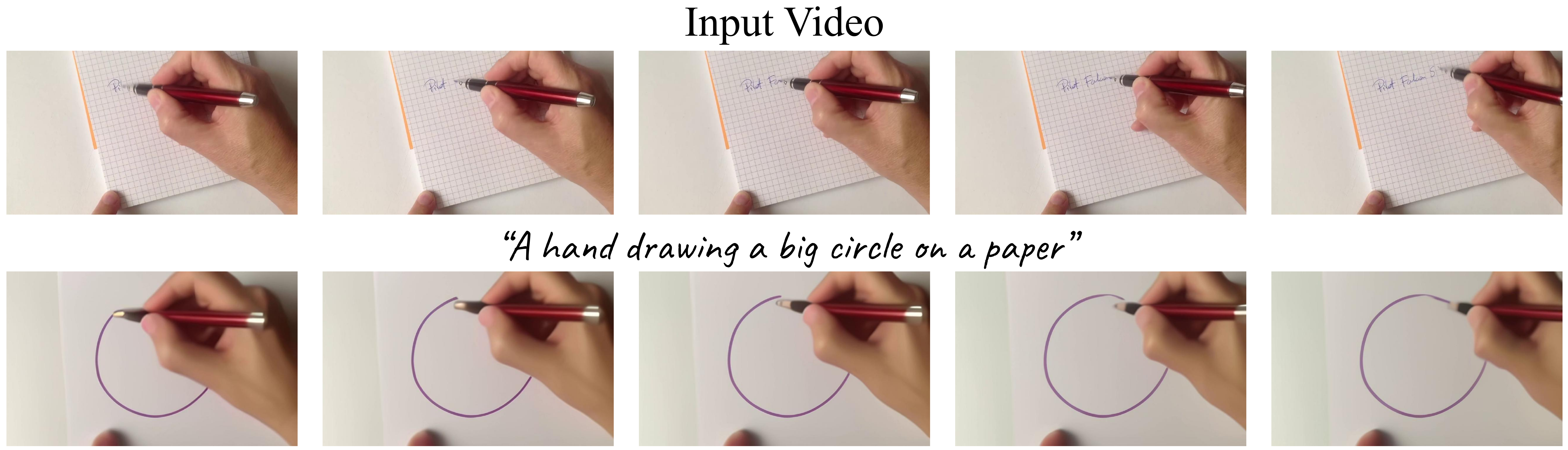} \\
\end{tabular}
 \caption{\textit{\textbf{Additional Video Editing Examples (1/4)}}}
\label{fig:sm_video_editing1}
\end{figure*}

\begin{figure*}[h!]
\begin{tabular}{c@{\hskip1pt}}

\includegraphics[width=0.98\linewidth]{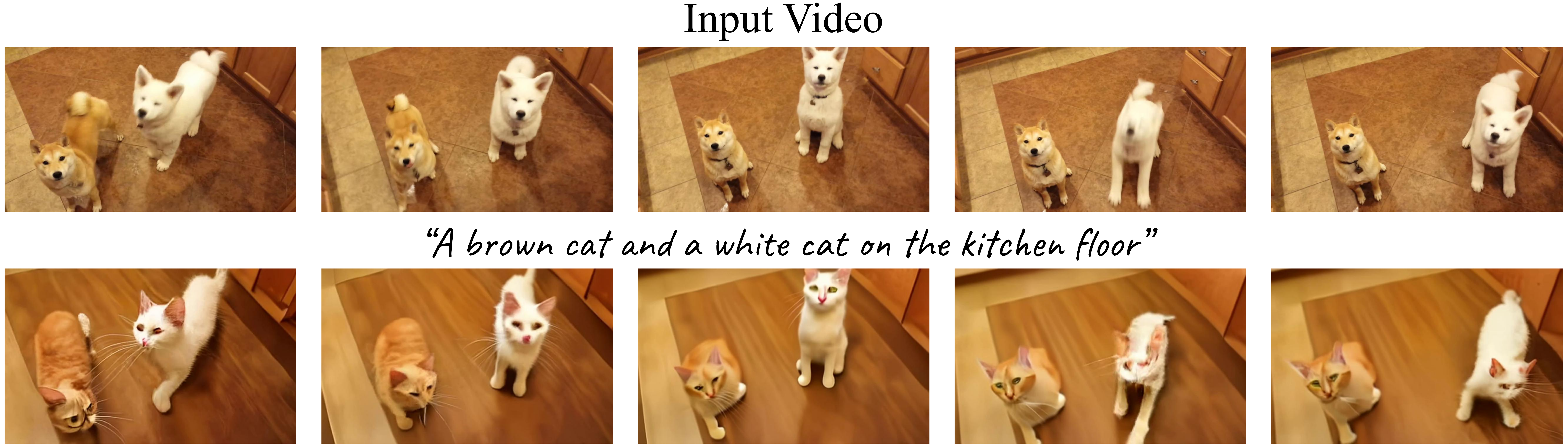} \\
\includegraphics[width=0.98\linewidth]{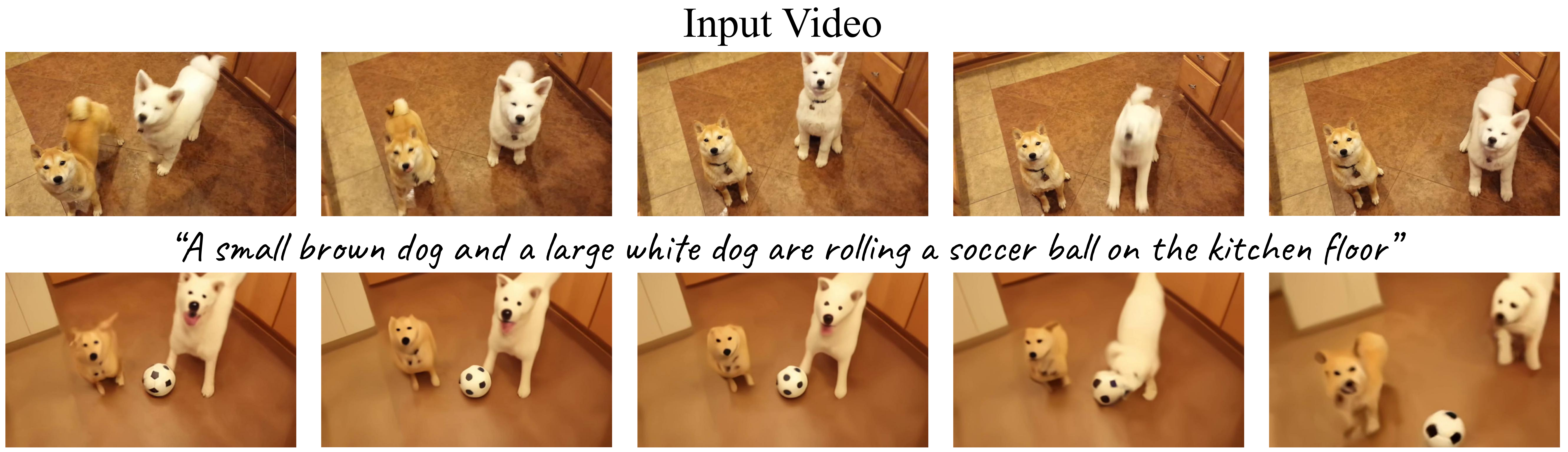} \\
\includegraphics[width=0.98\linewidth]{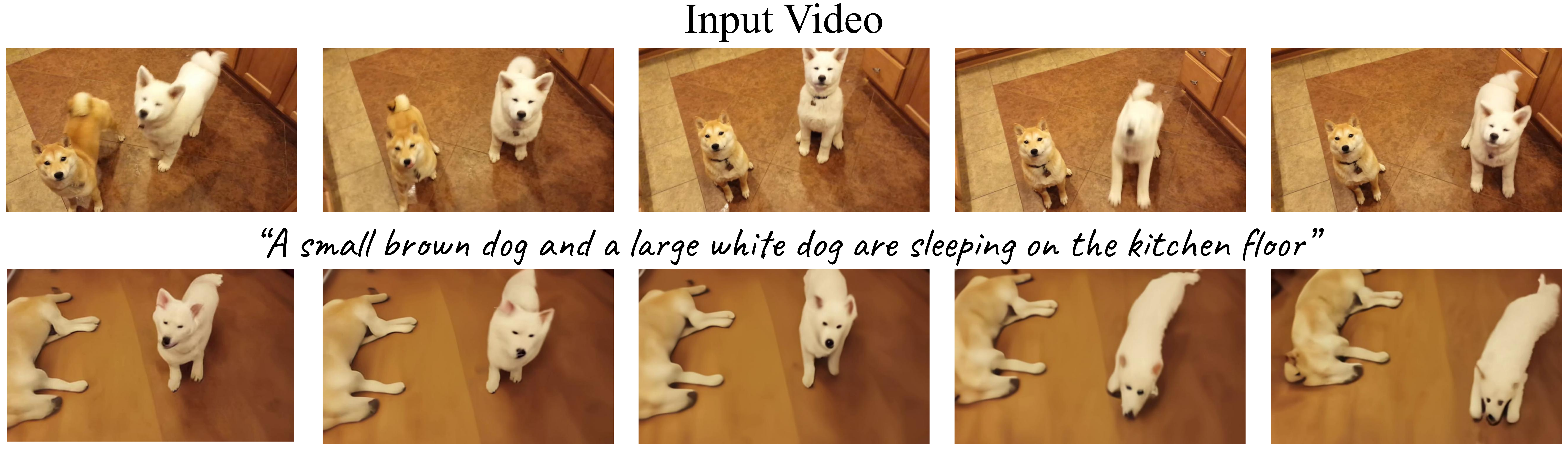} \\
\includegraphics[width=0.98\linewidth]{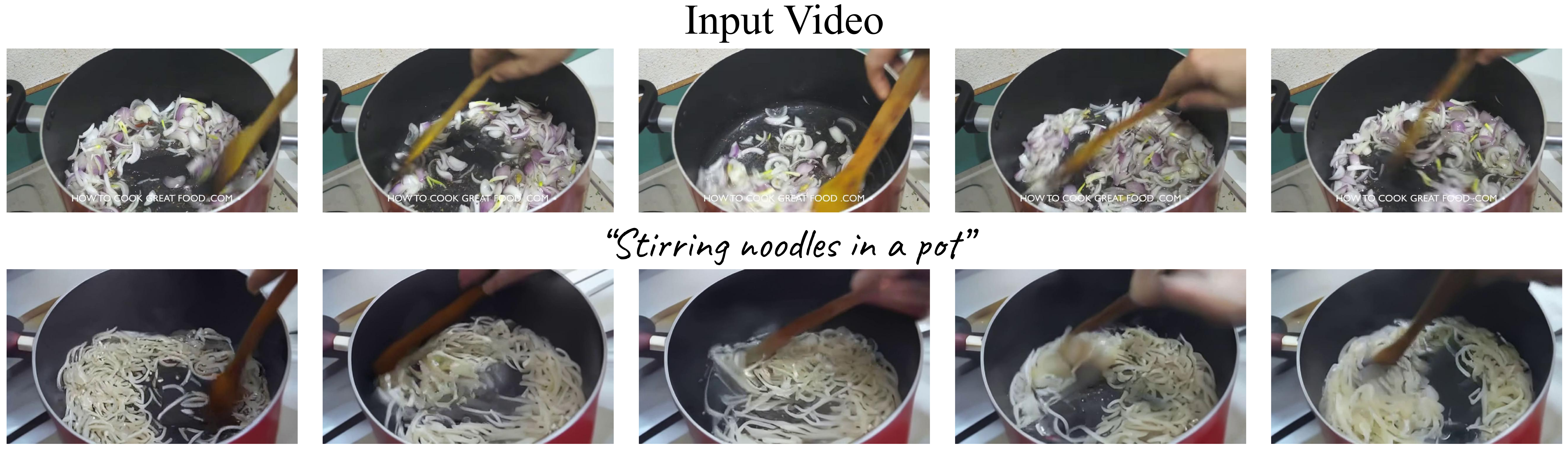} \\
\end{tabular}
 \caption{\textit{\textbf{Additional Video Editing Examples (2/4)}}}
\label{fig:sm_video_editing2}
\end{figure*}

\begin{figure*}[h!]
\begin{tabular}{c@{\hskip1pt}}
\includegraphics[width=0.98\linewidth]{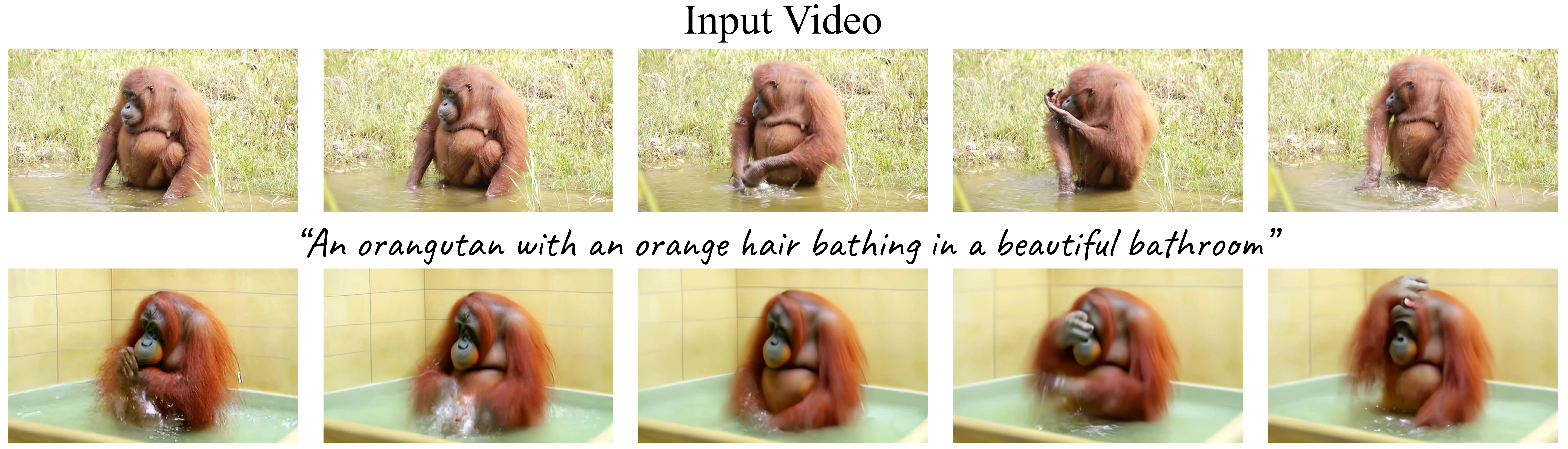} \\
\includegraphics[width=0.98\linewidth]{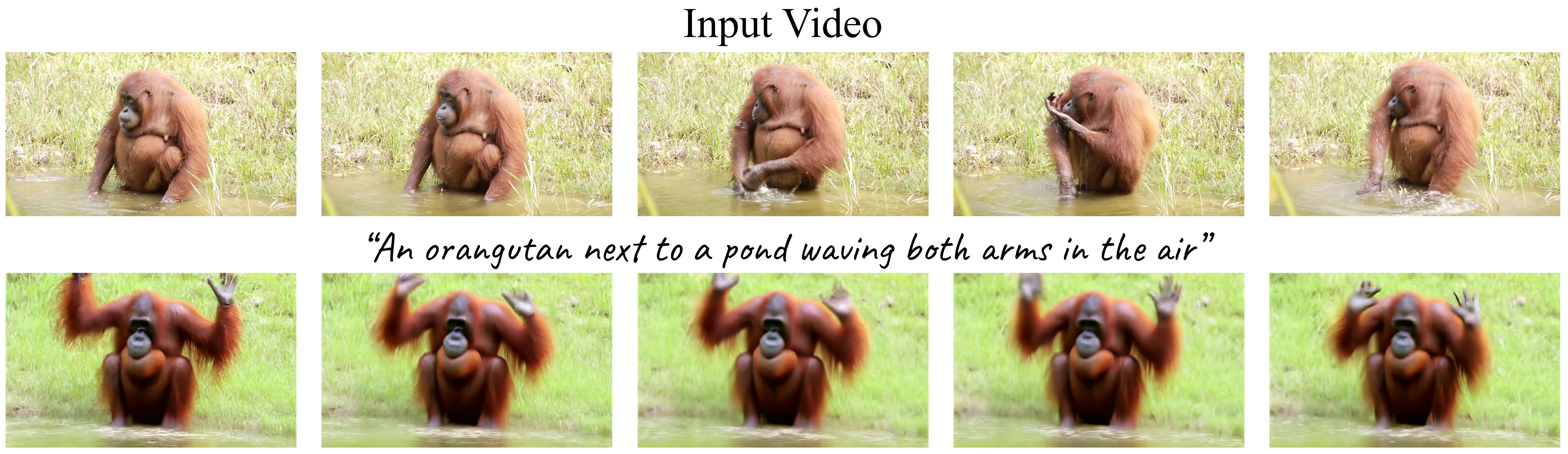} \\
\includegraphics[width=0.98\linewidth]{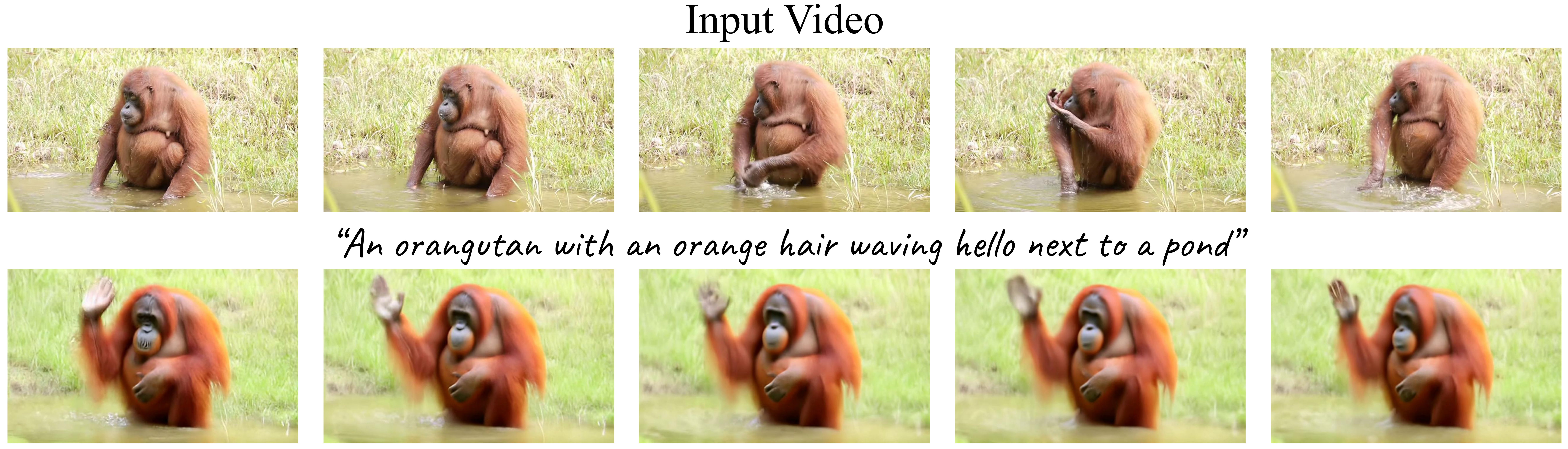} \\
\includegraphics[width=0.98\linewidth]{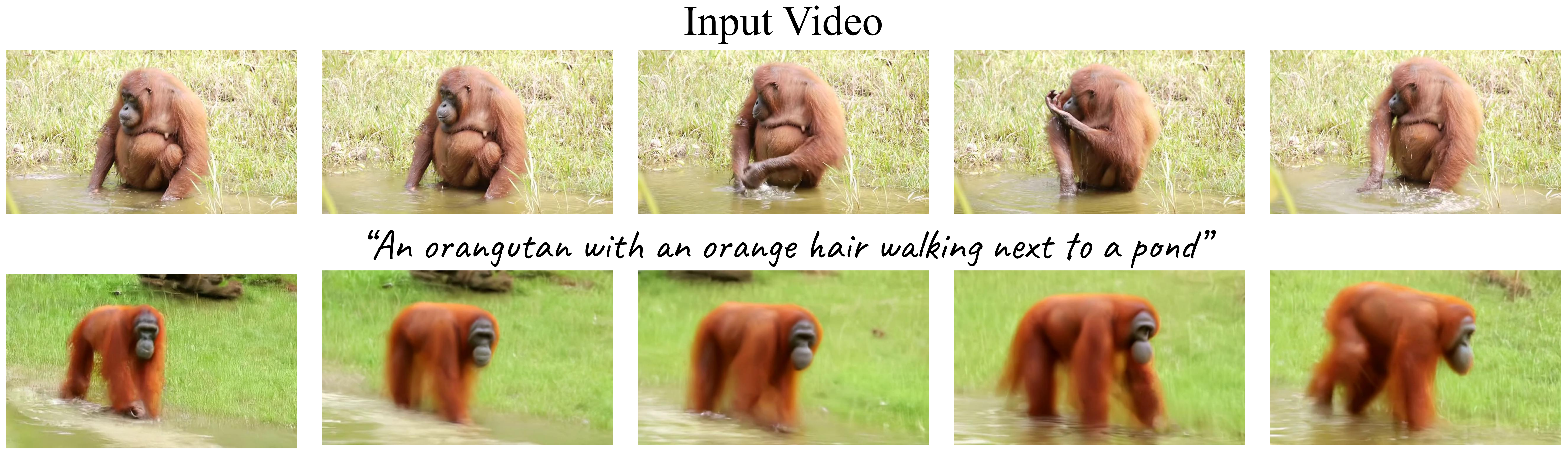} \\
\end{tabular}
 \caption{\textit{\textbf{Additional Video Editing Examples (3/4)}}}
\label{fig:sm_video_editing3}
\end{figure*}

\begin{figure*}[h!]
\begin{tabular}{c@{\hskip1pt}}
\includegraphics[width=0.98\linewidth]{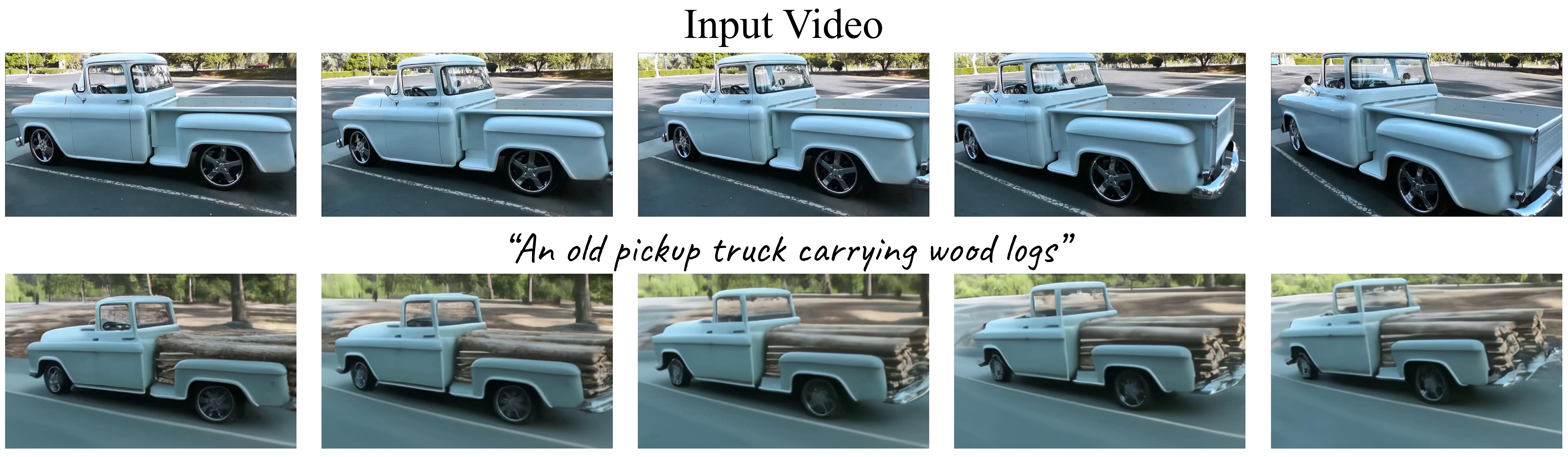} \\
\includegraphics[width=0.98\linewidth]{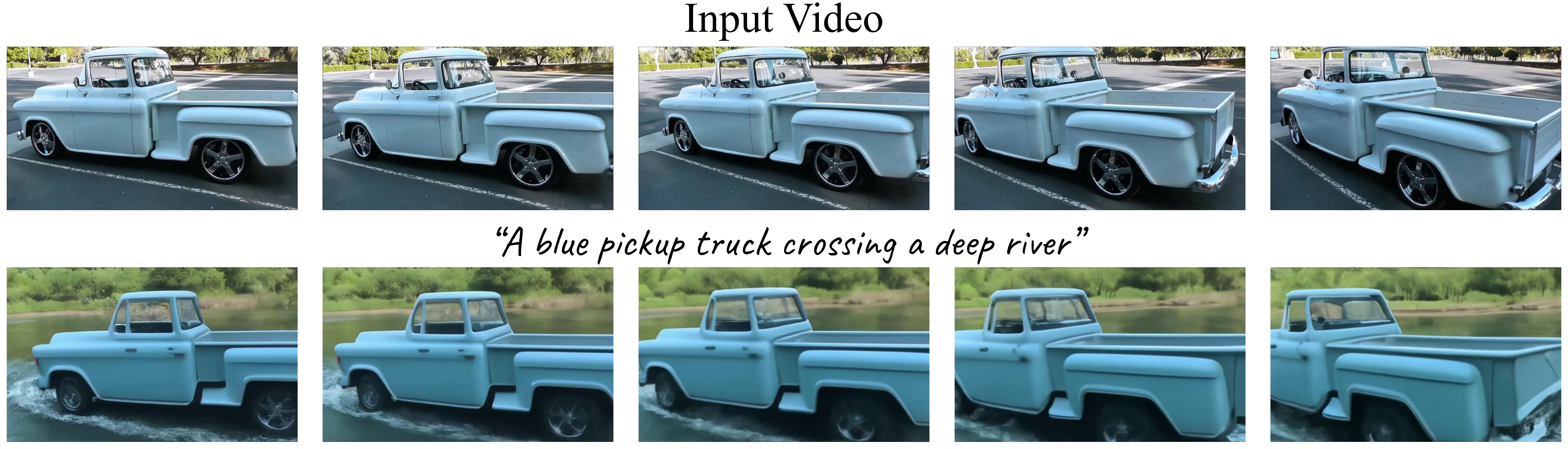} \\
\includegraphics[width=0.98\linewidth]{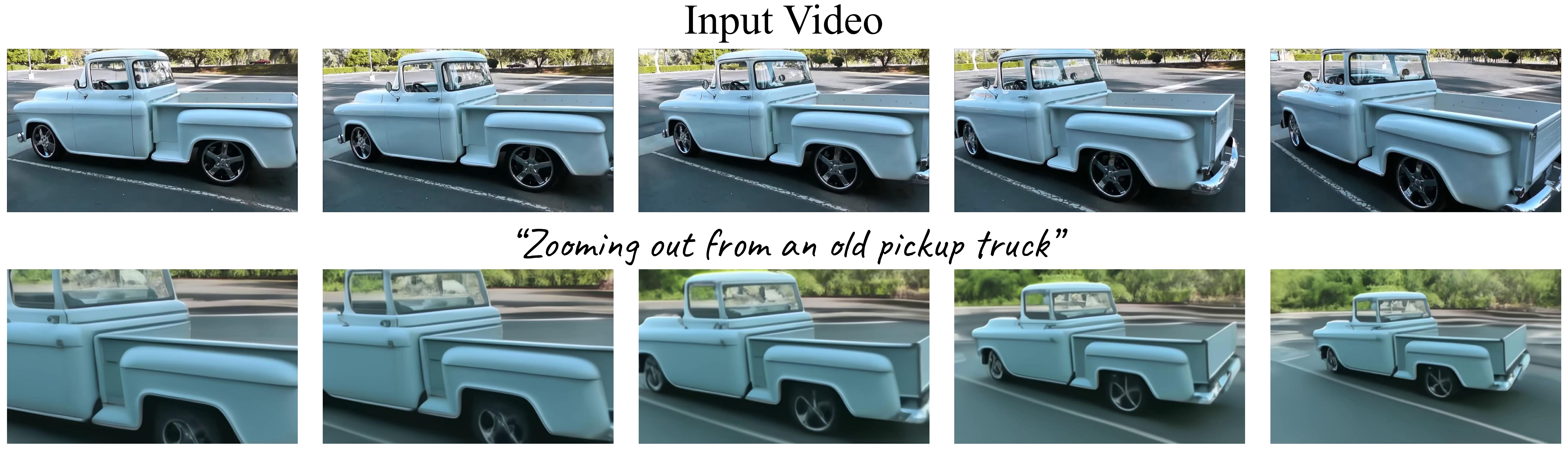} \\
\includegraphics[width=0.98\linewidth]{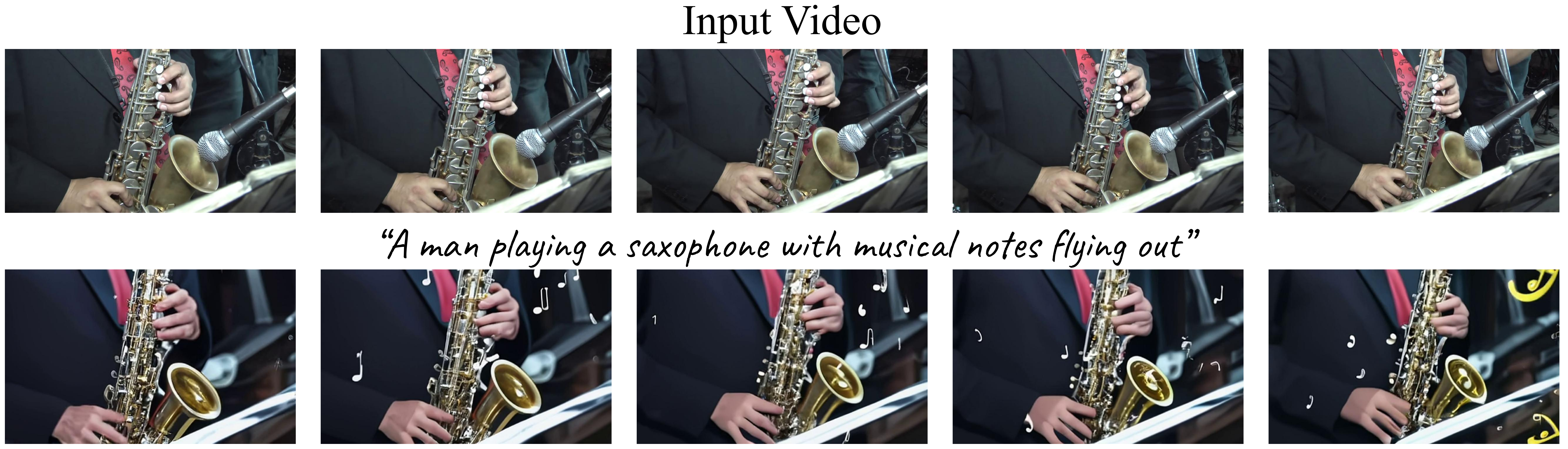} \\
\end{tabular}
 \caption{\textit{\textbf{Additional Video Editing Examples (4/4)}}}
\label{fig:sm_video_editing4}
\end{figure*}

\begin{figure*}[h!]
\begin{tabular}{c@{\hskip1pt}}
\includegraphics[width=0.98\linewidth]{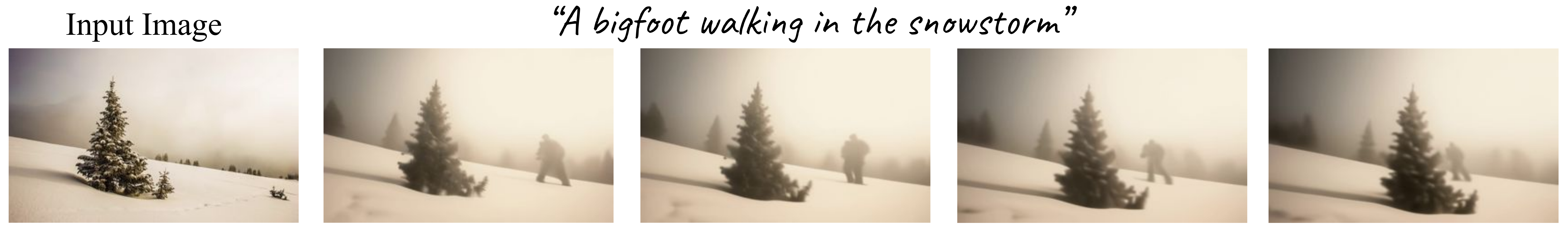} \\
\includegraphics[width=0.98\linewidth]{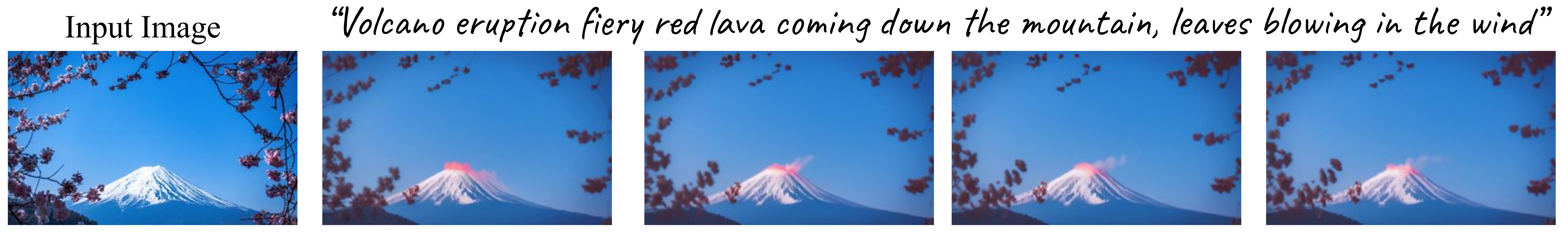} \\
\end{tabular}
 \caption{\textit{\textbf{Additional Image-to-Video Examples}}}
\label{fig:sm_img2vid}
\end{figure*}

\begin{figure*}[h!]
\begin{tabular}{c@{\hskip1pt}}
\includegraphics[width=0.98\linewidth]{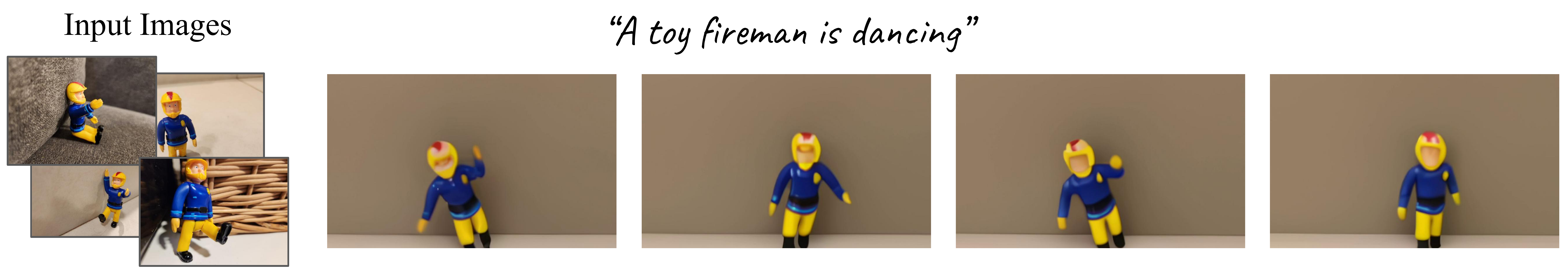} \\
\includegraphics[width=0.98\linewidth]{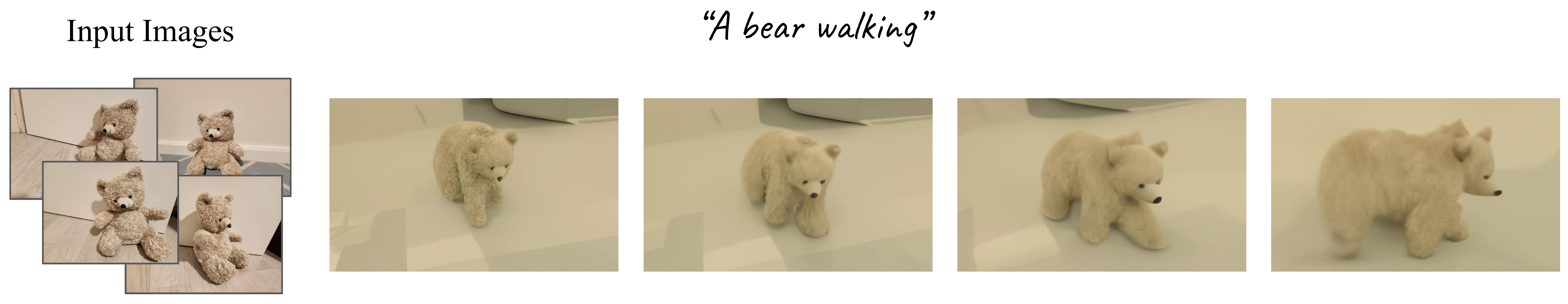} \\

\end{tabular}
 \caption{\textit{\textbf{Additional Subject-Driven Video Generation}}}
\label{fig:sm_dreambooth_video1}
\end{figure*}

\end{document}